\documentclass[10pt,twocolumn,letterpaper]{article}

\usepackage[algorithms]{wacv}      
\usepackage{graphicx}
\usepackage{booktabs}

\usepackage{algorithm}
\usepackage{algpseudocode}
\usepackage{amsmath}
\usepackage{multirow}
\usepackage{xcolor}
\usepackage{minted}
\usepackage{cuted}
\usepackage{caption}
\usepackage{soul}

\makeatletter
\let\old@maketitle\@maketitle
\renewcommand{\@maketitle}{%
  \old@maketitle%
  \begin{center}
    \includegraphics[width=1.0\linewidth,keepaspectratio]{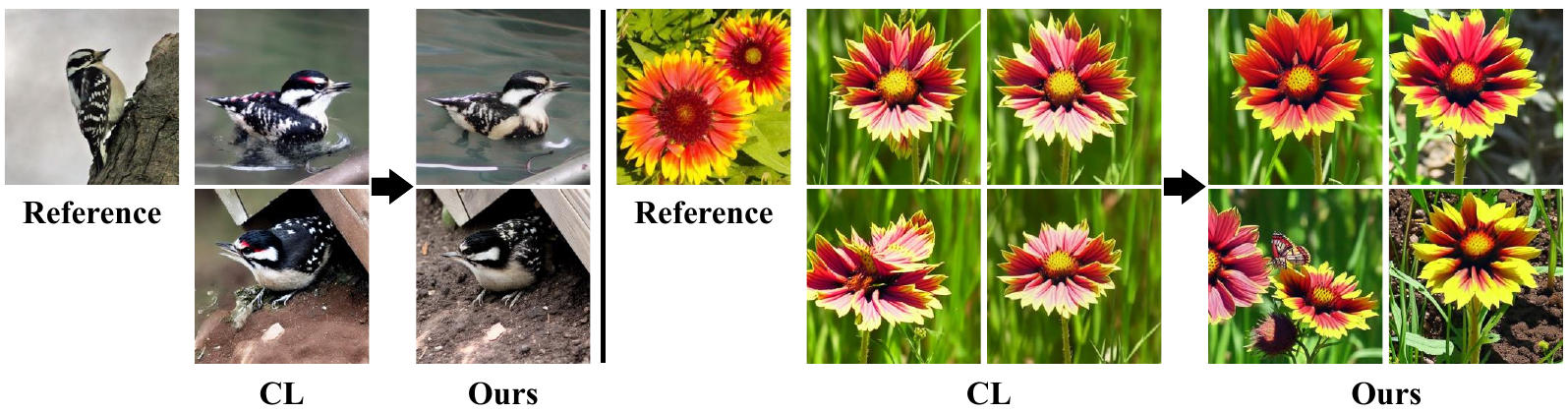}
    \captionof{figure}{
    Rare-class generations illustrating the two failure modes.
    (Left) Ours corrects foreground artifacts in concept-learning (CL) (\eg, incorrect red coloration on the bird’s head) and better preserves foreground details.
    (Right) CL generates similar 
    outputs despite varying prompts; Ours produces diverse prompt-conditioned samples while preserving foreground details.
    }
    \label{fig:teaser}
  \end{center}
  \vspace{1em}%
}
\makeatother

\usepackage[table]{xcolor}

\definecolor{bestcol}{RGB}{255, 230, 204}
\definecolor{secondcol}{RGB}{230, 247, 214}

\definecolor{cb_orange}{rgb}{0.9,0.49,0.0}
\definecolor{cb_blue}{rgb}{0.22,0.49,0.72}
\definecolor{cb_purple}{rgb}{0.49, 0.1, 0.89}
\definecolor{cb_green}{rgb}{0.25,0.67,0.22}
\definecolor{cb_red}{rgb}{0.89,0.1,0.11}
\definecolor{cb_pink}{rgb}{1, 0, 0.4}

\newcommand{\best}[1]{\cellcolor{bestcol}\textbf{#1}}
\newcommand{\second}[1]{\cellcolor{secondcol}#1}

\usepackage[accsupp]{axessibility}  

\definecolor{wacvblue}{rgb}{0.21,0.49,0.74}
\usepackage[pagebackref,breaklinks,colorlinks,allcolors=wacvblue]{hyperref}

\def\wacvPaperID{1008} 
\def\confName{WACV}
\def\confYear{2027}

\title{Improving Diffusion-Based Augmentation for Data-Scarce Image Classification via Mask-Guided Reference Attention and Repulsive Noise Optimization}

\author{
Hyun-Jic Oh$^{1,2}$\thanks{Equal contribution.}
\qquad
GaYeon Koh$^{1}$\footnotemark[1]
\qquad
Won-Ki Jeong$^{1,2}$\thanks{Corresponding author.}
\\[0.4em]
$^{1}$Korea University, Republic of Korea
\qquad
$^{2}$VIENCE Inc., Republic of Korea
\\
{\tt\small \{hyunjic0127,wkjeong\}@korea.ac.kr} \quad {\small Code: \url{https://github.com/hvcl/MGRAttn-RNO}}
}

\begin{document}
\maketitle

\begin{abstract}


Diffusion-based data augmentation has emerged as a powerful strategy for addressing data-scarce image classification. 
Recent pipelines typically adapt a pretrained text-to-image diffusion model to a limited number of class examples and subsequently sample synthetic images using diverse textual prompts. 
%
However, in extreme data-scarce regimes such as long-tailed single-sample settings, this approach suffers from a dual failure mode:
the generated samples often lack discriminative foreground details crucial for classifier training, and the adapted class prior can bias the early denoising process, causing output images to exhibit similar patterns despite varying text prompts.
%
To overcome these limitations, we propose a sampling-side framework of two complementary components that simultaneously enhance fine details and diversity.
First, Mask-Guided Reference (MGR) Attention updates only foreground value tokens with reference features via zero-shot segmentation, limiting reference influence to the target object and mitigating excessive global feature injection.
Second, Repulsive Noise Optimization (RNO) encourages initial noise trajectories to diverge from dominant generative modes of the adapted model, enabling target prompts to express broader contextual variation. 
Extensive evaluations show consistent improvements, with the largest gains in long-tailed single-sample settings of up to 2.50 and 2.94 points on CUB-LT IF100, and further robustness across few-shot and background-shift evaluations.

\end{abstract}


\section{Introduction}


Training high-performing image classifiers under data-scarce regimes remains a fundamental challenge, particularly in long-tailed and single-sample settings where certain categories contain only a handful of labeled instances~\cite{zhang2023deep, miyai2024generalized, xu2025deep}.
This constraint is especially severe in fine-grained classification, where deployable vision backbones must capture subtle, highly localized foreground attributes such as specific texture patterns or color contours to effectively distinguish visually similar sub-categories.


Synthetic data augmentation powered by pretrained text-to-image diffusion models~\cite{rombach2022high, nichol2021glide} has emerged as a promising solution to alleviate this scarcity by generating additional training samples with diverse synthetic variants~\cite{he2022synthetic,azizi2023synthetic,oh2024controllable}.
For these synthetic samples to benefit downstream classifier training, they must simultaneously satisfy two criteria: preserving class-discriminative fine-details while maintaining broad contextual diversity.



Diffusion-based augmentation methods typically adapt a text-to-image diffusion model via concept-learning techniques such as Textual Inversion~\cite{gal2023textual} or LoRA~\cite{hu2022lora}, and sample synthetic images using diverse textual prompts.
While effective to an extent, we observe that in extreme data-scarce regimes such as long-tailed single-sample cases, this approach suffers from a distinct dual failure mode (see~\cref{fig:teaser}).
(i) \textit{Fine-Detail Degradation}: the adapted model often fails to reproduce precise, discriminative foreground details, as adapting from only a few examples can yield weak fine-grained priors. 
(ii) \textit{Diversity Collapse}: the adapted class prior can dominate the early denoising process, causing generated images to collapse into similar geometric structures despite varying text prompt suffixes.
To overcome fine-detail degradation, we propose Mask-Guided Reference (MGR) Attention, a sampling-time localized attention strategy.
Existing attention-manipulation methods~\cite{cao2023masactrl,jeong2024visual} are designed for image editing such as style transfer or spatial layout control, where replacing self-attention key-value tokens with those from a reference is appropriate.
However, directly applying such techniques to data augmentation is not straightforward: global key-value replacement propagates reference features across the entire synthesized image, suppressing contextual variation driven by the text prompt.
Instead, MGR Attention uses a zero-shot segmenter~\cite{kirillov2023segment,carion2025sam} to obtain foreground boundaries, updating only foreground value tokens with reference features. 
To address diversity collapse, we introduce Repulsive Noise Optimization (RNO), a latent optimization method applied before the denoising process begins.
While standard noise optimization techniques like CNO~\cite{kim2025diverse} enforce an attraction-repulsion balance to stabilize generic generation, we observed that retaining an attraction term in adapted-model augmentation forces the latent trajectories right back into the dominant collapsed class prior. 
%
RNO establishes fixed anchors from the adaptation prompt, representing the dominant generative modes of the adapted model, and optimizes initial latent trajectories via a repulsion-only objective, encouraging diverse prompt-conditioned generation across broader layout spaces.
%



Our framework operates entirely at sampling time, making it compatible with existing concept-learning pipelines on UNet-based latent diffusion models as a plug-and-play component.
This is particularly effective in single-sample regimes, where concept-learning approaches cannot exploit reference information beyond what was captured during adaptation, whereas our modules leverage a real reference image at every generation step.
Extensive evaluations demonstrate consistent improvements in classification accuracy across long-tailed single-sample, few-shot, and Waterbirds background-shift settings, with the largest gains in the most data-scarce regimes.
Our main contributions are summarized as follows:
\begin{itemize}
    \item
    We introduce MGR Attention, a sampling-time localized attention strategy that updates target foreground value tokens with reference features via zero-shot segmentation, limiting reference influence to the target object region.
    
    \item
    We propose RNO, a latent optimization method that refines 
    initial noise trajectories via a repulsion-only objective against fixed anchors from the adaptation prompt, encouraging diverse prompt-conditioned generation.
   
    \item
    We demonstrate consistent improvements with the largest gains in long-tailed single-sample settings, and further robustness across few-shot and background-shift settings.
    
\end{itemize}


\label{sec:intro}

\section{Related Work}




\subsection{Diffusion Augmentation for Classification}

Generative data augmentation leveraging text-to-image diffusion models~\cite{rombach2022high, nichol2021glide,koh2025synthetic} has become a standard pipeline for expanding limited datasets.
A common research direction focuses on training-based adaptation strategies, where a diffusion model is adapted to a targeted dataset via concept-learning pipelines.
For instance, early formulations such as Da-Fusion~\cite{trabucco2024effective} fine-tune text embeddings via Textual Inversion~\cite{gal2023textual}, while subsequent frameworks like Diff-Mix~\cite{wang2024enhance} incorporate cross-class imagery blending.
Furthermore, recent adaptation lines like Diff-II~\cite{wang2025inversion} execute intra-class interpolation using target prompt suffixes, and more recent variants optimize this adaptation stage by learning salient concepts~\cite{zhao2025salient}, employing hierarchical structural guidance~\cite{lu2025higfa}, or enriching contextual prompt diversity during adaptation~\cite{yang2026beyond}.
While these training-based adaptation methods yield notable improvements for data distribution expansion, they encounter performance bottlenecks in extreme data-scarce scenarios such as single-sample settings, where interpolation-based strategies become ill-defined and foreground fidelity depends heavily on the adapted prior.

Our framework complements these adaptation-centric approaches by operating at sampling time without modifying any adapted model weights, and can therefore be combined with existing concept-learning baselines.

\subsection{Reference-Guided Diffusion Sampling}

A prominent branch of training-free diffusion manipulation modifies attention matrices to inject specific reference attributes.
Prompt-to-Prompt~\cite{hertz2023prompt} regulates semantic editing via cross-attention layout cloning, and Visual Style Prompting~\cite{jeong2024visual} replaces target key--value features to transplant global artistic aesthetics.
MasaCtrl~\cite{cao2023masactrl} converts self-attention into mutual self-attention for spatial transfer, using attention maps to derive localized masks for region-selective feature injection.
However, such attention-derived masks are less reliable in adapted diffusion models, as the learned class prior can distort intermediate attention maps and lead to inaccurate foreground localization.
These methods are primarily designed for image editing or style transfer, where maintaining a global spatial layout or uniform texture is the objective.
For data augmentation, global key-value token replacement causes reference features to propagate across the entire synthesized image, suppressing contextual variation that should be steered by the text prompt.

Another line of research focuses on region-controlled generation~\cite{xie2023boxdiff,chen2024anydoor} or detection-oriented augmentation frameworks~\cite{zhu2025recon,li2025domain}.
However, these methods require dense annotations or explicit object-level objectives, making them difficult to apply in automated data augmentation pipelines under image-level labels.

In contrast, MGR Attention operates under image-level labels only, employing a zero-shot segmenter~\cite{kirillov2023segment,carion2025sam} to automatically localize foreground boundaries and restricting reference feature insertion to foreground value tokens.

\subsection{Sampling-Time Diversity Control}


Controlling the generative trajectory at inference time without model retraining has been actively explored to enhance sample variance.
CADS~\cite{sadat2024cads} perturbs text conditioning signals to broaden output distributions.
InitNO~\cite{guo2024initno} refines initial noise maps for prompt-layout alignment, while CNO~\cite{kim2025diverse} optimizes initial noises in the Tweedie prediction space by balancing instance attraction and sample repulsion.
More recently, Harrington~\etal~\cite{harrington2026s} demonstrate that calibrating noise properties can recover diversity from repetitive sampling modes of pre-trained diffusion models.


Although successful in diverse sample generation, directly applying these techniques to adapted-model augmentation is not straightforward.
Perturbing conditioning signals can deteriorate sample fidelity in fine-grained settings where discriminative details are critical.
Furthermore, where the adapted class prior strongly biases early latent trajectories, retaining an attraction term makes it difficult for latent trajectories to escape the dominant generative mode.

RNO addresses these limitations by reformulating noise optimization for adapted diffusion models, treating adaptation-prompt anchors as stationary repulsive barriers and optimizing initial noises via a repulsion-only formulation.
Together, RNO and MGR Attention address the dual failure mode: RNO promotes diversity through noise optimization, while MGR Attention reinforces foreground details through localized reference guidance.

\begin{figure*}[t]
    \centering
    \includegraphics[width=1.0\linewidth]{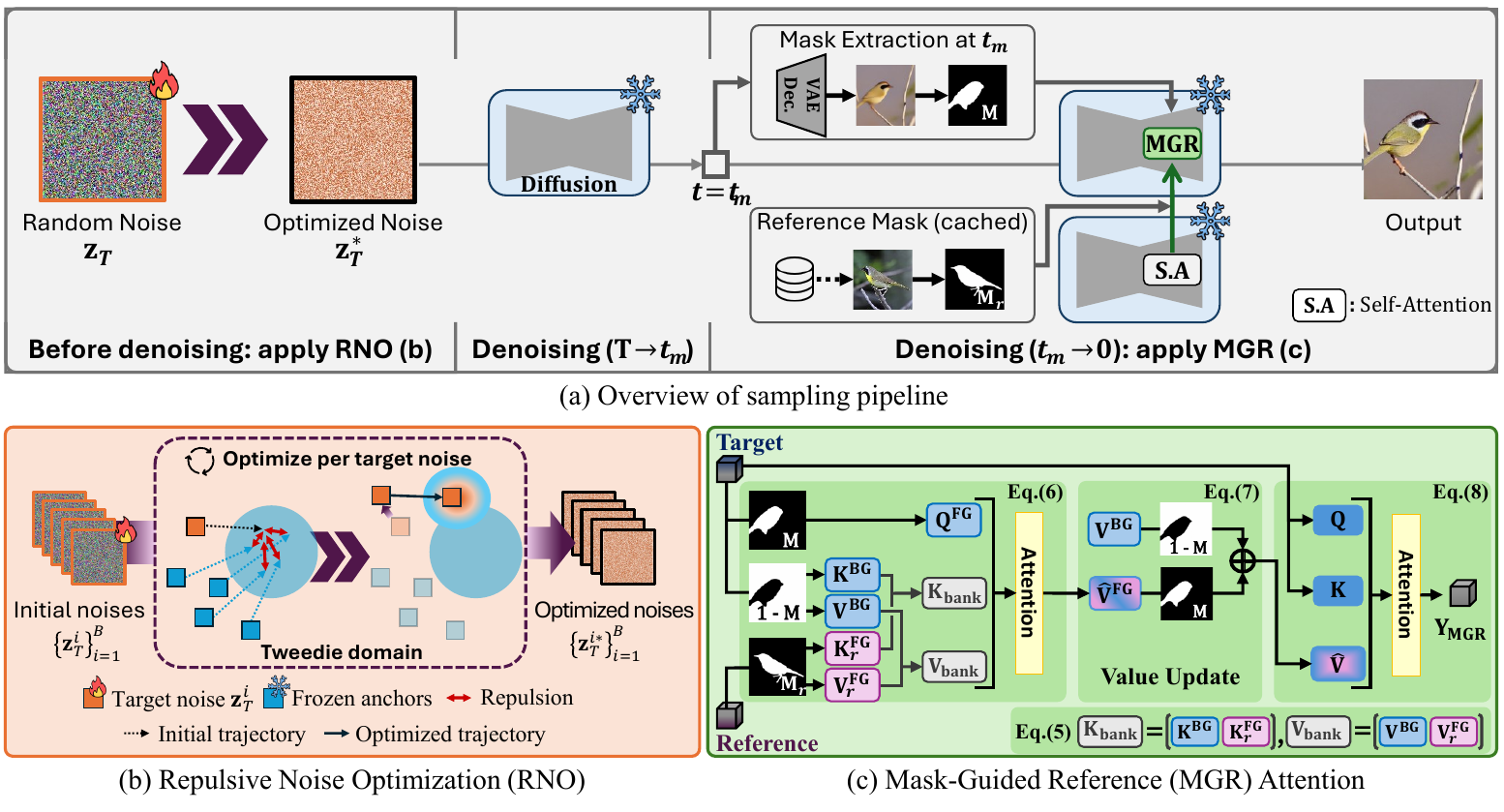}
    \caption{
    Overview of the proposed sampling pipeline.
    (a) RNO optimizes initial noises before denoising; the target mask $\mathbf{M}$ is extracted at $t_m$ and the reference mask $\mathbf{M}_r$ is cached offline; MGR Attention refines target foreground values during the remaining steps.
    (b) RNO optimizes each target noise ($\mathbf{z}_T^i$) to diverge from adaptation-prompt anchors (blue), which remain fixed throughout optimization.
    (c) MGR Attention constructs a hybrid bank $\mathbf{K}_\text{bank}, \mathbf{V}_\text{bank}$ from target-background and reference-foreground tokens, updating only foreground values. 
    }
    \label{fig:overview}
\end{figure*}

\label{sec:rel_work}

\section{Preliminaries}

\noindent\textbf{Latent Diffusion Process.}
Latent Diffusion Models (LDMs)~\cite{rombach2022high} perform the diffusion process within the lower-dimensional latent space of a pretrained autoencoder.
Let $\mathbf{z}_0$ denote the latent representation of a clean image.
The forward process gradually corrupts $\mathbf{z}_0$ over discrete timesteps $t \in \{0, \dots, T\}$:
\begin{equation}
    \mathbf{z}_t =
    \sqrt{\bar{\alpha}_t}\,\mathbf{z}_0 + \sqrt{1-\bar{\alpha}_t}\,\boldsymbol{\epsilon},
    \qquad
    \boldsymbol{\epsilon}\sim\mathcal{N}(\mathbf{0},\mathbf{I}),
    \label{eq:forward_process}
\end{equation}
where $\alpha_t = 1 - \beta_t$ and $\bar{\alpha}_t = \prod_{s=1}^t \alpha_s$ are determined by the predefined noise schedule.
In conditional generation, an adapted denoiser $\boldsymbol{\epsilon}_{\theta}(\mathbf{z}_t, t, \mathbf{c})$ predicts the noise under a text-conditioning embedding $\mathbf{c}$.
Following DDPM~\cite{ho2020denoising}, the Tweedie estimate of the clean latent at timestep $t$ is:
\begin{equation}
    \hat{\mathbf{z}}_{0|t}(\mathbf{z}_t, \mathbf{c}) = \frac{1}{\sqrt{\bar{\alpha}_t}} \left( \mathbf{z}_t - \sqrt{1-\bar{\alpha}_t}\,\boldsymbol{\epsilon}_{\theta}(\mathbf{z}_t, t, \mathbf{c}) \right).
    \label{eq:tweedie_estimate}
\end{equation}
The forward process in~\cref{eq:forward_process} is used to noise the reference image in MGR Attention; the Tweedie estimate in~\cref{eq:tweedie_estimate} forms the basis of our noise optimization in RNO.

\noindent\textbf{Self-Attention in Latent Diffusion.}
At each self-attention layer in the LDMs, an intermediate feature tensor is projected into query ($\mathbf{Q}$), key ($\mathbf{K}$), and value ($\mathbf{V}$) tokens, and the output is computed as:
\begin{equation}
    \operatorname{Attn}(\mathbf{Q},\mathbf{K},\mathbf{V})
    =
    \operatorname{Softmax}
    \left(
    \frac{\mathbf{Q}\mathbf{K}^{\top}}{\sqrt{d}}
    \right)
    \mathbf{V},
    \label{eq:self_attention}
\end{equation}
where $d$ is the feature channel dimension.
Our method modifies this token update rule without altering any pre-trained or adapted parameter weights.
As the attention weights are determined by $\mathbf{Q}$ and $\mathbf{K}$, our method exploits this property by modifying only selected value tokens, confining direct reference influence to the attended features without altering the attention weight distribution.
%



\label{sec:preliminaries}

\section{Method}

\subsection{Problem Setup}
\label{sec:setup}

We assume a text-to-image diffusion model has been adapted to the available training data via concept-learning pipelines, establishing an adapted class prior.
We refer to the prompt used during this adaptation as the adaptation prompt (\eg, ``a photo of a \texttt{<concept>} bird'').

During augmentation, a real training image from the target class is selected as the reference image. 
Each generated sample is conditioned on a text prompt describing the desired contextual variation, such as background or pose.



\subsection{Pipeline Overview}
\label{sec:overview}
Our framework operates entirely during inference and consists of two complementary components (see~\cref{fig:overview}a).
Before the diffusion loop begins ($t{=}T$), 
RNO optimizes a set of initial latent noises for diverse sample generation.
The denoising then proceeds to an intermediate timestep $t_m$, at which a foreground mask is extracted from the decoded intermediate image via zero-shot segmentation.
During the remaining steps ($t_m{\to}0$), MGR Attention performs a region-selective value update in selected self-attention layers, infusing reference features into the target object region.

\subsection{Repulsive Noise Optimization (RNO)}
\label{sec:rno}

When adapted from extremely limited data, the learned prior can heavily bias early latent trajectories toward repetitive structural modes.
CNO~\cite{kim2025diverse} further struggles here, as the dominant prior can overwhelm the attraction objective, pulling trajectories back to repetitive generation.

To address this, RNO introduces a repulsion-only latent optimization strategy before the denoising process begins (see~\cref{fig:overview}b). 
%
Let $\{\mathbf{z}_T^i\}_{i=1}^B$ be a batch of initial random noises assigned to target variation prompts, and let $\mathbf{c}_t^i$ and $\mathbf{c}_a$ denote the embeddings of the target and adaptation prompts, respectively. 
The remaining $B{-}1$ branches conditioned on $\mathbf{c}_a$ serve as adaptation-prompt anchors for each target noise $\mathbf{z}_T^i$, kept frozen throughout optimization via $\operatorname{sg}[\cdot]$.
These anchors represent dominant generative modes of the adapted model, serving as stationary repulsive barriers.
Let $\hat{\mathbf{z}}^i {=} \hat{\mathbf{z}}_{0|T} (\mathbf{z}_T^i, \mathbf{c}_t^i)$ and $\hat{\mathbf{z}}^j_a {=} \hat{\mathbf{z}}_{0|T} (\mathbf{z}_T^j, \mathbf{c}_a)$ denote target and anchor predictions.
RNO minimizes their average similarity at timestep $T$:
\begin{equation}
    \mathcal{L}_{\mathrm{RNO}} = \frac{1}{B(B{-}1)} 
    \sum_{i=1}^B \sum_{j \neq i} 
    \operatorname{sim} \left( 
    \hat{\mathbf{z}}^i,\, 
    \operatorname{sg}[\hat{\mathbf{z}}^j_a]
    \right),
    \label{eq:rno_loss}
\end{equation}
where $\operatorname{sim}(\cdot, \cdot)$ denotes cosine similarity and $\operatorname{sg}[\cdot]$ the stop-gradient operation.
This repulsion-only formulation encourages the optimized latents $\{\mathbf{z}_T^{i*}\}_{i=1}^B$ to explore broader layout spaces for diverse sample generation. 

\subsection{Mask-Guided Reference (MGR) Attention}
\label{sec:mgr}

While RNO promotes layout diversity, generated objects can still lack fine-grained details; MGR Attention addresses this via a region-selective value update that injects reference features into foreground token positions (see~\cref{fig:overview}c).

\subsubsection{Mask Extraction}
\label{sec:ext_masks}

At timestep $t_m$, where coarse object structure has emerged, we decode the Tweedie estimate $\hat{\mathbf{z}}_{0|t_m}$ and apply a zero-shot segmenter to extract target foreground mask $\mathbf{M}$; reference mask $\mathbf{M}_r$ is pre-computed from the clean reference image offline.
Both masks are downscaled to match the spatial resolution of the selected self-attention layers.

\begin{figure*}[t!]
    \centering
    \includegraphics[width=\linewidth]{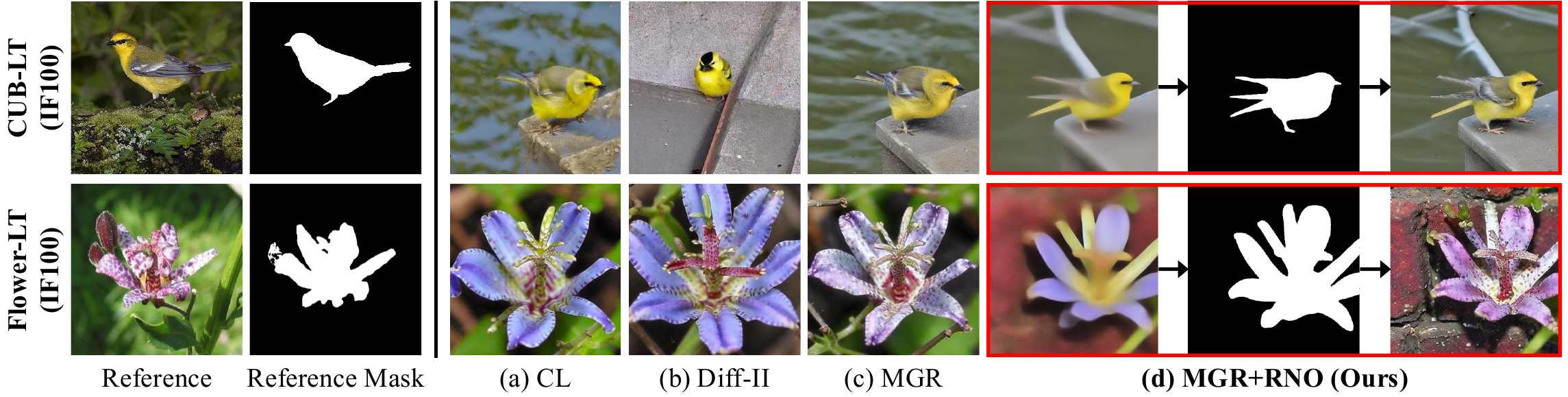}
    \caption{
    Qualitative comparison on CUB-LT and Flower-LT IF100 (prompts: ``sitting on a ledge by water'' and ``against a brick wall'').
    (a) CL and (b) Diff-II rely on the adapted class prior, limiting foreground details and prompt-conditioned variation.
    (c) MGR enhances foreground details.
    (d) MGR+RNO (Ours) increases prompt-conditioned diversity while preserving foreground details, successfully following the given prompts.
    }
    \label{fig:qual_comparison}
\end{figure*}

\subsubsection{Region-Selective Value Update}
\label{sec:mgr_update}

MGR Attention operates on two concurrent streams: a target stream generating the synthetic image and a frozen reference stream supplying reference features.
At each timestep $t {\leq} t_m$, the reference image is encoded and noised via~\cref{eq:forward_process}; this noised latent is passed through the frozen denoiser to extract key and value tokens from the selected self-attention layers, serving solely as a feature provider.

Let $\mathbf{Q}, \mathbf{K}, \mathbf{V}$ and $\mathbf{K}_r, \mathbf{V}_r$ denote tokens of the target and reference streams, respectively, partitioned by spatial masks into foreground ($\Omega^{\mathrm{FG}}$) and background ($\Omega^{\mathrm{BG}}$) sets.
%
MGR Attention constructs a hybrid bank from target-background and reference-foreground tokens (see~\cref{fig:overview}c):
\begin{equation}
    \mathbf{K}_{\mathrm{bank}} = [\mathbf{K}^{\mathrm{BG}}; \mathbf{K}_r^{\mathrm{FG}}], \qquad \mathbf{V}_{\mathrm{bank}} = [\mathbf{V}^{\mathrm{BG}}; \mathbf{V}_r^{\mathrm{FG}}].
    \label{eq:mgr_bank}
\end{equation}
Foreground queries attend exclusively to this bank to compute reference-fortified values:
\begin{equation}
    \hat{\mathbf{V}}^{\mathrm{FG}} = \operatorname{Attn}(\mathbf{Q}^{\mathrm{FG}}, \mathbf{K}_{\mathrm{bank}}, \mathbf{V}_{\mathrm{bank}}).
    \label{eq:mgr_fg}
\end{equation}
The value update is then applied to foreground indices:
\begin{equation}
    \hat{\mathbf{V}}[\Omega^{\mathrm{FG}}] = \hat{\mathbf{V}}^{\mathrm{FG}}, \qquad \hat{\mathbf{V}}[\Omega^{\mathrm{BG}}] = \mathbf{V}[\Omega^{\mathrm{BG}}].
    \label{eq:mgr_merge}
\end{equation}
Finally, the self-attention output is computed using the updated value tensor $\hat{\mathbf{V}}$:
\begin{equation}
    \mathbf{Y}_{\mathrm{MGR}} = \operatorname{Attn}(\mathbf{Q}, \mathbf{K}, \hat{\mathbf{V}}).
    \label{eq:mgr_output}
\end{equation}
Since $\mathbf{Q}$ and $\mathbf{K}$ remain directly unaffected, $\mathbf{Y}_{\mathrm{MGR}}$ preserves the original attention weight distribution, confining reference influence to the foreground value tokens.
\label{sec:method}

\section{Experiment}

\subsection{Setup}
\label{sec:exp_setup}

\noindent\textbf{Benchmarks.}
We evaluate in three data-scarce regimes:
(i) \emph{Long-tailed (LT)}: CUB-LT~\cite{samuel2021generalized} and Flower-LT~\cite{wang2024enhance} with imbalance factors $\mathrm{IF}{\in}\{100,20,10\}$, following Diff-II~\cite{wang2025inversion}.
(ii) \emph{Few-shot}: 5-shot and 10-shot on CUB~\cite{wah2011caltech}, Aircraft~\cite{maji2013fine}, Cars~\cite{krause20133d}, and Pet~\cite{parkhi2012cats}.
(iii) \emph{Background-shift}: 5-shot CUB trained model evaluated on Waterbirds~\cite{sagawa2019distributionally}.


\noindent\textbf{Generation and adaptation.}
We use a pretrained Stable Diffusion (SD)~\cite{rombach2022high} v1.5 model with 50-step DDIM sampling~\cite{song2020denoising} and a classifier-free guidance scale of 7.0, with publicly available adapted checkpoints from Diff-Mix~\cite{wang2024enhance} and class-specific prompt suffixes following Diff-II~\cite{wang2025inversion}.
We additionally evaluate on top of BOB~\cite{yang2026beyond} on Flower-LT, reproduced following the official protocol, with both SD v1.5 and SD v2.1 backbones.
MGR Attention extracts the target mask at $t_m{=}30$ and is applied to decoder up-block self-attention layers (from layer 2.1) for $t{\leq}t_m$, using SAM3~\cite{carion2025sam} for zero-shot segmentation.
Further details, including RNO hyperparameters, are provided in the supplementary material.

\begin{table*}[t]
\centering
\caption{
LT classification results (\%) on CUB-LT and Flower-LT ($\Delta$CE: gain over CE; Diff-II$\dagger$: SYNAuG~\cite{ye2023synaug}-style normalization).
}
\resizebox{\linewidth}{!}{
\begin{tabular}{
l|cccc|cccc|cccc|cccc}
\toprule 
& \multicolumn{4}{c|}{CUB-LT (ResNet-50)}
& \multicolumn{4}{c|}{Flower-LT (ResNet-50)}
& \multicolumn{4}{c|}{CUB-LT (ViT-B/16)}
& \multicolumn{4}{c}{Flower-LT (ViT-B/16)} \\
Method 
& IF100 & IF20 & IF10 & $\Delta$CE 
& IF100 & IF20 & IF10 & $\Delta$CE 
& IF100 & IF20 & IF10 & $\Delta$CE 
& IF100 & IF20 & IF10 & $\Delta$CE \\
\midrule 
CE
    & 33.65 & 44.82 & 58.13 & - 
    & 80.43 & 90.87 & 95.07 & - 
    & 53.87 & 68.36 & 78.82 & -    
    & 92.84 & 97.52 & 99.20   & -     
    \\
Mixup~\cite{zhang2017mixup}
    &   37.59   &  49.15   &  63.43   &  +4.52    
    &   81.11  &  92.13   &  96.16   &  +1.01    
    &   53.64  &  68.85   &  79.38   &  +0.27    
    &   92.06  &  97.79   &  98.96   &  -0.25    
    \\
CutMix~\cite{yun2019cutmix}
    &   37.61  &  49.15   &  63.27   &  +4.48    
    &   81.96  &  92.28   &  95.80   &  +1.22    
    &   54.02  &  68.45   &  78.22   &  -0.12    
    &   92.06  &  97.28   &  99.00   &  -0.41    
    \\
CMO~\cite{park2022majority}
    & 32.94 & 44.08 & 57.62 & -0.65
    & 83.95 & 91.43 & 95.19 & +1.40
    & 49.22  &  64.20   &  79.55 & -2.69    
    &  92.21 &  97.57  &  99.03  & -0.25  
    \\
CMO-DRW~\cite{cao2019learning}
    & 32.57 & 46.43 & 59.25 & +0.55
    & 84.07 & 92.02 & 95.29 & +1.67
    &  51.20 &  67.29 & 73.80 &  -2.92    
    &  95.27 &  98.15 & 99.17 &  +1.01    
    \\
Real-Filter~\cite{he2022synthetic}
    &  36.83  &  49.41  & 62.69 &  +4.11    
    &  80.38  &  91.58  & 95.65 &  +0.41    
    &  48.84  &  64.77  & 78.37 &  -3.02    
    &  88.30  &  96.16  & 98.88 &  -2.07    
    \\
Real-Guidance~\cite{he2022synthetic}   
    &  36.50  &  49.34  &  59.11   &  +2.78    
    &  81.94  &  91.60  &  95.27   &  +0.81    
    &  52.62  &  65.43  &  75.37   &  -2.54    
    &  91.16  &  96.75  &  98.69    &  -0.99    
    \\
Real-Gen~\cite{wang2024enhance}
    & 45.86 & 53.43 & 61.42 & +8.04  
    & 83.56 & 91.84 & 95.22 & +1.42  
    & 61.17 & 69.97 & 76.46  &  +2.18    
    & 92.67 & 96.63 & 99.00  &  -0.42    
    \\
Real-Mix~\cite{wang2024enhance}
    & 47.75 & 55.67 & 62.27 & +9.70  
    & 85.19 & 92.96 & 96.02 & +2.60  
    & 62.72 & 71.75 & 76.35 &  +3.26    
    & 92.60 & 96.12 & 98.30 &  -0.85    
    \\
Da-Fusion~\cite{trabucco2024effective}
    & 35.74 & 46.82 & 57.13 &  +1.03    
    & 83.86 & 91.82 & 95.31 &  +1.54    
    & 49.13 & 62.19 & 72.80  &  -5.64    
    & 92.13 & 96.92 & 98.71  &  -0.60    
    \\
Diff-AUG~\cite{wang2024enhance}
    & 47.77 & 59.73 & 67.22 & +12.71     
    & 87.47 & 94.12 & 96.87 & +4.03     
    & 54.02 & 68.76 & 80.82 & +0.85        
    & \second{\underline{95.61}} & 97.57 & 99.22 & +0.95        
    \\
Diff-Mix~\cite{wang2024enhance}
    & 50.35 & 58.19 & 64.48 & +12.14 
    & 89.46 & 93.86 & 96.63 & +4.53  
    & 60.89 & 66.33 & 73.08  & -0.25    
    & 95.29 & 97.77 & 99.00  & +0.83   
    \\
Diff-II~\cite{wang2025inversion}
    & 51.21 & 62.31 & \second{\underline{70.28}} & +15.74 
    & \second{\underline{89.54}} & 94.39 & \second{\underline{97.35}} & \second{\underline{+4.97}}  
    & 64.50 & 74.53  & 81.46  & +6.48    
    & 95.34 & 98.52 & \second{\underline{99.37}}   & \second{\underline{+1.22}}     
    \\
Diff-II$\dagger$~\cite{wang2025inversion}
    & \second{\underline{53.59}} & \second{\underline{64.14}} & 69.52 & \second{\underline{+16.89}}   
    & 88.44 & \second{\underline{94.49}} & 96.84 & +4.47  
    & \second{\underline{66.93}} &  \second{\underline{75.54}} &  \second{\underline{81.72}}  &  \second{\underline{+7.71}}  
    & 92.81 &  \second{\underline{98.57}} &  \best{99.44}  &  +0.42    
    \\ 
\hline 
\textbf{Ours}
    & \best{56.09} & \best{66.15} & \best{71.68} & \best{+19.11} 
    & \best{90.34} & \best{95.29} & \best{97.60} & \best{+5.62}  
    & \best{69.87} & \best{77.51} & \best{82.59} & \best{+9.64}    
    & \best{96.13} & \best{98.59} & \best{99.44} & \best{+1.53}     
    \\
\toprule 
\end{tabular}
}
\label{tab:lt_results}
\end{table*}

\noindent\textbf{Synthetic-data and classifier protocols.}
For LT experiments, we generate enough samples to match each class to the maximum real class frequency, following SYNAuG~\cite{ye2023synaug}.
For few-shot and background-shift experiments, real and synthetic images are mixed at a 1:5 ratio following Diff-II~\cite{wang2025inversion}.
We train ResNet-50~\cite{he2016deep} and ViT-B/16~\cite{dosovitskiy2020vit} classifiers for all augmentation methods.


\noindent\textbf{Baselines and metrics.}
We compare with standard cross entropy (CE), Mixup~\cite{zhang2017mixup}, CutMix~\cite{yun2019cutmix}, CMO~\cite{park2022majority}, CMO-DRW~\cite{cao2019learning}, and diffusion-based augmentation methods including Real-Filter, Real-Guidance~\cite{he2022synthetic}, Real-Gen, Real-Mix, Diff-AUG, Diff-Mix~\cite{wang2024enhance}, Da-Fusion~\cite{trabucco2024effective}, and Diff-II~\cite{wang2025inversion}.
Top-1 accuracy (\%) is the primary metric.



%
%

\subsection{Qualitative Comparison}
\Cref{fig:qual_comparison} compares generated samples on CUB- and Flower-LT IF100.
On CUB-LT, (a) CL and (b) Diff-II fail to reproduce fine-grained foreground details such as wing texture and head color, whereas (c) MGR and (d) MGR+RNO better preserve these foreground details while generating diverse backgrounds.
On Flower-LT, (a) and (b) drift from the reference color patterns and fail to follow the given prompt (\eg, ``against a brick wall'');
(d) MGR+RNO successfully generates prompt-conditioned samples while preserving the distinctive color and petal structure of the reference flower.

\subsection{Downstream Classification}


\noindent\textbf{Long-tailed.}
%
\Cref{tab:lt_results} reports results on CUB- and Flower-LT across three imbalance factors and two classifier architectures, where our method achieves the best accuracy in all reported settings.
The improvement is largest in the most data-scarce CUB-LT IF100 setting: relative to Diff-II${\dagger}$ under the same class-normalized synthetic-data protocol, our method improves ResNet-50 accuracy from 53.59\% to 56.09\% ($+2.50$ points) and ViT-B/16 accuracy from 66.93\% to 69.87\% ($+2.94$ points).
On Flower-LT, gains are smaller where strongest baselines already exceed 94--99\% accuracy, but our method remains competitive.

\begin{table}[t!]
\centering
\setlength{\tabcolsep}{2pt}
\caption{
Few-shot (5-shot and 10-shot) classification results. 
}
\resizebox{\linewidth}{!}{
\begin{tabular}{l|ccccc|ccccc}
    \toprule 
    & \multicolumn{5}{c|}{\textbf{5-shot}} & \multicolumn{5}{c}{\textbf{10-shot}} \\
    Method & CUB & Air. & Pet & Car & Avg & CUB & Air. & Pet & Car & Avg \\ 
    \midrule 
    CE
        & 54.52 & 36.63 & 86.98 & 40.10 & 54.56
        & 70.26 & 54.11 & 90.41 & 69.64 & 71.11 \\
    Mixup
        & 54.33 & 36.75 & 86.36 & 41.43 & 54.72
        & 70.00 & 55.28 & 89.49 & 70.28 & 71.26 \\
    CutMix
        & 51.74 & 34.20 & 85.41 & 41.89 & 53.31
        & 68.91 & 51.05 & 89.71 & 67.71 & 69.35 \\
    Real-Filter
        & 55.03 & 40.40 & 87.66 & 57.60 & 60.17
        & 67.36 & 47.21 & 90.14 & 69.64 & 68.59 \\
    Real-Guidance
        & 54.67 & 36.27 & 87.22 & 41.60 & 54.90
        & 69.81 & 54.77 & 90.00 & 70.97 & 71.40 \\
    Real-Mix
        & 51.00 & 32.21 & 86.42 & 53.56 & 55.80
        & 64.65 & 44.16 & 89.00 & 75.72 & 68.38 \\
    Da-Fusion
        & 59.40 & 34.98 & 87.41 & 58.73 & 60.13
        & 72.05 & 51.14 & 90.47 & 77.61 & 72.82 \\
    Diff-AUG
        & 61.14 & 39.42 & 89.23 & 54.08 & 60.97
        & 71.94 & 53.04 & 91.00 & 75.66 & 72.91 \\
    Diff-Mix
        & 56.18 & 32.61 & 88.77 & 56.39 & 58.49
        & 70.16 & 52.33 & 89.78 & 72.39 & 71.17 \\
    Diff-II
        & 62.22 & \textbf{42.15} & \textbf{89.84} & 64.24 & 64.61
        & 72.66 & 57.43 & 91.01 & 82.02 & 75.78 \\
    \midrule
    \textbf{Ours}
        & \textbf{66.98} & 41.04 & 89.67 & \textbf{67.49} & \textbf{66.30}
        & \textbf{74.11} & \textbf{57.67} & \textbf{91.25} & \textbf{83.58} & \textbf{76.65} \\
    \bottomrule \specialrule{0.1pt}{0pt}{0pt}
\end{tabular}
}
\label{tab:fewshot_results}
\end{table}

\noindent\textbf{Few-shot.}
%
\Cref{tab:fewshot_results} summarizes 5-shot and 10-shot results on CUB, Aircraft, Pet, and Cars, where our method achieves the highest average accuracy in both regimes, outperforming Diff-II by 1.69 and 0.87 points respectively.
The gains are driven by CUB ($+4.76$ points) and Car ($+3.25$ points) in the 5-shot setting; 
on Aircraft, where large intra-class appearance variation makes reference anchoring less effective, Diff-II remains slightly better ($-1.11$ points), while the 10-shot setting shows improvements on all four datasets.

\begin{table}[t!]
\centering
\renewcommand{\arraystretch}{0.8}
\caption{
Background-shift evaluation on Waterbirds.
}
\setlength{\tabcolsep}{3pt}
\resizebox{\linewidth}{!}{
\begin{tabular}{lccccc}
    \toprule 
    Method & (L,L) & (W,W) & (L,W) & (W,L) & Avg \\ 
    \midrule 
    CE 
        & 38.55 & 32.15 & 37.90 & 30.20 & 34.70 \\
    Real-Filter~\cite{he2022synthetic}
        & 38.87 & 34.72 & 38.25 & 31.64 & 35.87 \\
    Real-Guidance~\cite{he2022synthetic}
        & 39.49 & 33.64 & 39.80 & 29.78 & 35.68 \\
    Real-Mix~\cite{wang2024enhance}
        & 30.29 & 29.44 & 30.02 & 26.95 & 29.18 \\
    Da-Fusion~\cite{trabucco2024effective}
        & 44.50 & 39.04 & 44.46 & 33.49 & 40.37 \\
    Diff-AUG~\cite{wang2024enhance}
        & 49.02 & 40.59 & 48.71 & 38.58 & 44.23 \\
    Diff-Mix~\cite{wang2024enhance}
        & 33.84 & 30.53 & 34.72 & 28.19 & 31.82 \\
    Diff-II~\cite{wang2025inversion}
        & 49.84 & 41.46 & 51.01 & 42.03 & 46.09 \\
    \midrule 
    \textbf{Ours}
        & \textbf{52.73} & \textbf{41.90} & \textbf{52.20} & \textbf{42.37} & \textbf{47.30} \\
    \bottomrule 
\end{tabular}
}
\label{tab:OOD_results}
\end{table}


\begin{figure*}[t!]
    \centering
    \includegraphics[width=.9\linewidth]{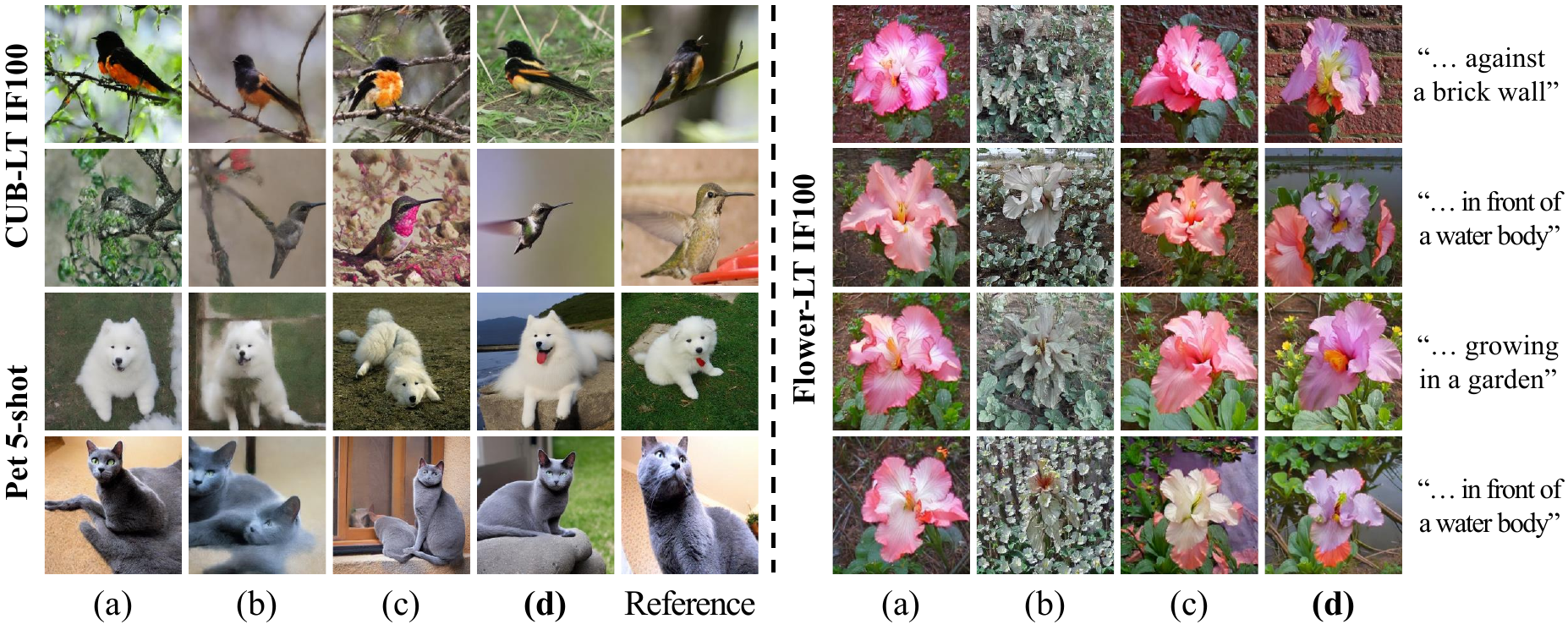}
    \caption{
    Qualitative module comparisons.
    (left) On CUB-LT IF100 and Pet 5-shot, (a) global K/V replacement (VSP-style~\cite{jeong2024visual}), (b) attention-masked K/V replacement (MasaCtrl~\cite{cao2023masactrl}-style), and (c) SAM3-masked K/V replacement introduce unwanted reference style or disrupt target layout, whereas (d) Ours better preserves target layout while enhancing localized foreground details.
    (right) On a single-image Flower-LT class with varying prompts, (a) CL collapses to similar outputs, (b) CADS increases variation at the cost of quality degradation, and (c) CNO generates diverse samples but struggles to follow prompt-conditioned variation, whereas (d) RNO produces diverse prompt-conditioned layouts while retaining recognizable foreground structure.
    }
    \label{fig:qual_module}
\end{figure*}

\noindent\textbf{Background-shift robustness.}
%
\Cref{tab:OOD_results} shows improvements in mean group accuracy from 46.09\% to 47.30\% ($+1.21$) and worst-group from 41.46\% to 41.90\% ($+0.44$).
This is consistent with our design: MGR Attention strengthens foreground cues, while RNO diversifies contexts in the training data, suggesting reduced spurious correlations between bird labels and background appearance.

\subsection{Ablation Studies}
\label{sec:ablation}
\noindent\textbf{Qualitative design analysis.}
\Cref{fig:qual_module} provides visual evidence for the design choices analyzed below.
For MGR Attention (left), global key--value replacement (K/V swap; VSP-style~\cite{jeong2024visual}) transfers reference style across the entire image; applying an attention-derived mask (MasaCtrl~\cite{cao2023masactrl}-style) reduces global leakage but remains imprecise in adapted diffusion models.
Using a K/V swap with SAM3 mask improves localization but can still over-transfer reference appearance.
MGR Attention with foreground-selective value update best preserves target layout while enhancing localized object details.
For RNO (right), CL and CNO remain close to repetitive class-prior modes, whereas CADS degrades sample quality; RNO produces broader prompt-conditioned layouts with recognizable foreground structure.

\noindent\textbf{Module Contributions.}
\Cref{tab:component_ablation} reports module ablations (CUB- and Flower-LT IF100).
MGR Attention improves Few and Single accuracy over CL on both datasets (CUB-LT: $+$1.75/$+$1.67; Flower-LT: $+$3.30/$+$4.75), confirming that foreground-selective value update strengthens discriminative details in data-scarce classes.
RNO alone shows mixed results: it improves CUB-LT ($+$0.33 overall) but slightly reduces Flower-LT ($-$0.58), indicating that sampling variation alone is insufficient without foreground guidance.
MGR+RNO achieves the best accuracy on both datasets, with gains in Few and Single groups ($+$3.13/$+$3.26 on CUB-LT; $+$4.81/$+$7.65 on Flower-LT), confirming complementary roles: RNO broadens sampled configurations while MGR reinforces foreground evidence.

\begin{table}[t]
\centering
\caption{
Module ablation on CUB-LT and Flower-LT IF100.
Single: 1 image per class (CUB: 83 classes, Flower: 37 classes).
Few: $<$5/$<$10 images per class, including Single.
}
\renewcommand{\arraystretch}{1.05}
\label{tab:component_ablation}
\resizebox{\columnwidth}{!}{
\begin{tabular}{lccc|ccc}
\toprule
& \multicolumn{3}{c|}{CUB-LT IF100} 
& \multicolumn{3}{c}{Flower-LT IF100} \\
Method & Few & Single & Avg 
       & Few & Single & Avg \\
\midrule
CL      & 42.95 & 37.79 & 54.16 
        & 77.37 & 64.39 & 87.74 \\
MGR     & 44.70 & 39.46 & 55.61 
        & 80.67 & 69.14 & 89.34 \\
RNO     & 43.23 & 37.18 & 54.49 
        & 76.35 & 61.37 & 87.16 \\
\textbf{MGR+RNO} 
        & \textbf{46.08} & \textbf{41.05} 
        & \textbf{56.09}
        & \textbf{82.18} & \textbf{72.04} 
        & \textbf{90.34} \\
\bottomrule
\end{tabular}
}
\end{table}

\begin{table}[t]
\centering
\caption{
Design choice ablation on CUB-LT and Flower-LT IF100.
(i) Attention mechanism. (ii) Noise optimization.
}
\label{tab:design_choices}
\renewcommand{\arraystretch}{1.05}
\resizebox{\columnwidth}{!}{

\begin{tabular}{llccc|ccc}
\toprule
& & \multicolumn{3}{c|}{CUB-LT IF100} 
  & \multicolumn{3}{c}{Flower-LT IF100} \\
& Method & Few & Single & Avg 
         & Few & Single & Avg \\
\midrule
\multirow{6}{*}{(i)}
& CL
        & 42.95 & 37.79 & 54.16
        & 77.37 & 64.39 & 87.74 \\
& K/V swap
        & 43.26 & 37.46 & 54.71
        & 79.41 & 66.36 & 88.57 \\
& K/V swap (Attn. mask)
        & 35.47 & 27.97 & 49.28
        & 79.12 & 66.13 & 88.66 \\
& K/V swap (SAM3)       
        & 51.26 & 31.67 & 51.36
        & 79.21 & 66.01 & 88.64 \\
& MGR (Attn. mask)
        & 43.70 & 37.71 & 54.71
        & 77.95 & 64.97 & 87.98 \\
& MGR (SAM3, Ours)      
        & 44.70 & 39.46 & 55.61 
        & 80.67 & 69.14 & 89.34 \\
\midrule
\multirow{2}{*}{(ii)}
& MGR+CNO~\cite{kim2025diverse}
        & 45.84 & 40.18 & 55.83 
        & 72.09 & 66.82 & 89.05 \\
& MGR+RNO (Ours)        
        & \textbf{46.08} & \textbf{41.05} 
        & \textbf{56.09}
        & \textbf{82.18} & \textbf{72.04} 
        & \textbf{90.34} \\
\bottomrule
\end{tabular}


}
\end{table}

\noindent\textbf{Design Choices.}
\Cref{tab:design_choices} ablates design choices on CUB- and Flower-LT IF100.
%
\noindent\textbf{\textit{(i) Attention mechanism.}}
On CUB-LT, global K/V swap slightly improves Few accuracy but degrades Single-group performance (37.46\% vs. 37.79\%); applying an attention-derived mask (MasaCtrl-style) further hurts both groups (Single: 27.97\%), confirming that attention-derived masks are unreliable in adapted diffusion models.
K/V swap with SAM3 mask improves Few (51.26\%) but substantially reduces Single accuracy (31.67\%), showing that key modification disrupts the adapted class structure even with accurate masks.
MGR's value-only update resolves this: replacing the attention-derived mask with SAM3 improves Single from 37.71\% to 39.46\%.
\noindent\textbf{\textit{(ii) Noise optimization.}}
Comparing CNO and RNO on top of MGR, MGR+RNO consistently outperforms MGR+CNO on Single-group accuracy (41.05\% vs.\ 40.18\% on CUB-LT; 72.04\% vs.\ 66.82\% on Flower-LT),
demonstrating that repulsion-only optimization better suits adapted diffusion models.

\begin{table}[t]
\centering
\renewcommand{\arraystretch}{0.9}
\caption{
Diversity and foreground detail metrics on single-sample classes of CUB-LT and Flower-LT IF100.
MSS~\cite{sadat2024cads} and Vendi~\cite{friedman2023vendi} measure diversity; Ref-FG Sim measures foreground similarity via DINOv2~\cite{oquab2023dinov2} features on SAM3~\cite{carion2025sam}-segmented regions.
Bold indicates best among MGR-based methods; CADS is excluded as it produces severe quality degradation (see~\cref{fig:qual_module}).
}
\resizebox{\columnwidth}{!}{
\begin{tabular}{lccc|ccc}
\toprule
       & \multicolumn{3}{c|}{CUB-LT IF100} 
       & \multicolumn{3}{c}{Flower-LT IF100} \\
Method & MSS$\downarrow$ & Vendi$\uparrow$ 
       & Ref-FG$\uparrow$ 
       & MSS$\downarrow$ & Vendi$\uparrow$ 
       & Ref-FG$\uparrow$ \\
\midrule
CL
    & 0.62 & 7.46 & 0.45 & 0.81 & 3.25 & 0.59 \\
CADS 
    & 0.35 & 17.51 & 0.22 & 0.56 & 8.89 & 0.28 \\
CNO
    & 0.58 & 8.39 & 0.44 & 0.76 & 4.12 & 0.56 \\
\midrule
MGR
    & 0.60 & 7.89 & 0.46 & 0.79 & 3.60 & \textbf{0.61} \\
MGR+CNO
    & 0.63 & 7.18 & 0.48 & 0.77 & 3.86 & 0.59 \\
MGR+RNO
    & \textbf{0.58} & \textbf{8.28} 
    & \textbf{0.49} & \textbf{0.72} 
    & \textbf{4.91} & 0.56 \\
\bottomrule
\end{tabular}
}
\label{tab:metric_line_analysis}
\end{table}


\noindent\textbf{Diversity--Detail Analysis.}
\Cref{tab:metric_line_analysis} analyzes diversity and foreground detail of synthetic samples by each method on single-sample classes of CUB-LT and Flower-LT IF100 (83 and 37 classes, respectively; 50 images per class).
We report MSS~\cite{sadat2024cads} and Vendi~\cite{friedman2023vendi} as diversity metrics, and Ref-FG Sim as a measure of foreground similarity to the reference image.
CADS achieves the highest diversity but substantially reduces foreground details (see also~\cref{fig:qual_module}).
CNO improves diversity over CL without improving Ref-FG Sim, while MGR consistently increases Ref-FG Sim, confirming that foreground-selective value update transfers discriminative reference details.
Although the absolute gain is modest, it translates to meaningful downstream accuracy improvements in Few and Single groups (\cref{tab:component_ablation}).
MGR+RNO (Ours) outperforms MGR+CNO in diversity on both datasets while achieving the highest Ref-FG Sim on CUB-LT; on Flower-LT, Ref-FG Sim is slightly lower than CL as RNO introduces greater layout variation, yet MGR+RNO attains the best downstream accuracy across all settings.
These results support that improving both diversity and foreground details contributes to downstream gains.

\begin{table}[t]
\centering
\renewcommand{\arraystretch}{.9}
\caption{
Mask extraction timestep analysis on CUB-LT and Flower-LT IF100.
}
\resizebox{\linewidth}{!}{
\begin{tabular}{lccc|ccc}
\toprule
& \multicolumn{3}{c|}{CUB-LT IF100} 
& \multicolumn{3}{c}{Flower-LT IF100} \\
$t_m$ & Few & Single & Avg 
      & Few & Single & Avg \\
\midrule
40  & 45.17 & 39.91 & 55.92 
    & 80.09 & 68.10 & 89.10 \\
\textbf{30~\checkmark}
    & \textbf{46.08} & \textbf{41.05} 
    & \textbf{56.09}
    & \textbf{82.18} & \textbf{72.04} 
    & \textbf{90.34} \\
20  & 44.72 & 39.50 & 55.58 
    & 80.52 & 67.75 & 89.27 \\
10  & 43.15 & 37.99 & 54.30 
    & 79.65 & 67.40 & 88.86 \\
\bottomrule
\end{tabular}
}
\label{tab:sensitivity}
\end{table}

\noindent\textbf{Mask extraction timestep.}
\Cref{tab:sensitivity} ablates the mask extraction timestep $t_m$ on CUB-LT and Flower-LT IF100.
Both datasets peak at $t_m{=}30$. Flower-LT shows higher sensitivity, with a larger performance drop at $t_m{=}10$ (Single: 67.40 vs.\ 72.04).
At $t_m{=}40$, masks may be extracted before sufficient object structure has emerged, likely limiting the effectiveness of reference-guided refinement, particularly on Flower-LT where larger foreground objects require more denoising steps to form reliable boundaries.
At $t_m{=}20/10$, the remaining denoising steps may be insufficient for MGR Attention to effectively incorporate reference features.
MGR Attention also shows robust performance across different zero-shot segmenters (see supplementary).

\begin{table}[t]
\centering
\caption{
Compatibility with BOB~\cite{yang2026beyond} on Flower-LT IF100. 
Reproduced following the official repository.
}
\renewcommand{\arraystretch}{.9}
\label{tab:bob_compatibility}
\resizebox{\linewidth}{!}{









\begin{tabular}{lccc|ccc}
\toprule
& \multicolumn{3}{c|}{SD v1.5} 
& \multicolumn{3}{c}{SD v2.1} \\
Method & Few & Single & Avg 
       & Few & Single & Avg \\
\midrule
BOB      
        & 79.70 & 69.58 & 85.49 
        & 80.52 & 70.27 & 86.03 \\
BOB+Ours 
        & \textbf{81.80} & \textbf{72.53} 
        & \textbf{86.80}
        & \textbf{83.14} & \textbf{75.64} 
        & \textbf{87.83} \\
\bottomrule
\end{tabular}

}
\end{table}


\noindent\textbf{Compatibility with alternative adaptation.}
\Cref{tab:bob_compatibility} applies our modules to BOB~\cite{yang2026beyond} on Flower-LT IF100, reproduced following its diffusion-adaptation and classifier-training protocols.
BOB+Ours consistently improves over BOB across both SD v1.5 and SD v2.1 backbones, with the gains in Single-group accuracy of 2.95 and 5.37 points.
These results show that, on Flower-LT, our sampling-time modules remain effective on top of an alternative concept-learning method.

\label{sec:exp}

\section{Conclusion}

We introduced a sampling-time diffusion augmentation framework combining MGR Attention and RNO for data-scarce image classification.
MGR Attention updates only foreground value tokens with reference features via zero-shot segmentation, strengthening fine-grained object details while limiting unwanted reference propagation.
RNO optimizes initial latent trajectories via a repulsion-only objective against adaptation-prompt anchors, promoting prompt-conditioned diversity.
Across long-tailed, few-shot, and background-shift evaluations, the proposed framework improves downstream classifier accuracy with the largest gains in Few and Single groups, confirming the complementary roles of the two modules.

\noindent\textbf{Limitations and future work.}
Applying a strong repulsion objective in RNO may push latent trajectories away from class-consistent regions.
A theoretical understanding of how repulsion strength affects the trade-off between diversity and class consistency remains an interesting direction for future work.
Beyond sampling-time interventions, incorporating reference-guided objectives into the concept-learning process itself could yield stronger adapted priors under extreme data scarcity, offering a complementary direction to our approach.
In addition, our experiments are limited to UNet-based Stable Diffusion backbones (v1.5 and v2.1); extending MGR Attention to transformer-based architectures such as MM-DiT, where image and text tokens share joint attention, is left for future work.
\section*{Acknowledgments}
This work was supported in part by the National Research Foundation of Korea under Grant RS-2024-00349697 and Grant RS-2021-NR060143; in part by the Technology Development Program (RS-2024-00437796) funded by the Ministry of SMEs and Startups (MSS, Korea); in part by the Institute for Information and Communications Technology Planning and Evaluation under Grant IITP-2027-RS-2020-II201819; and in part by Korea University Grant.

{
    \small
    \bibliographystyle{ieeenat_fullname}
    \bibliography{main}

@String(CVPR= {IEEE Conf. Comput. Vis. Pattern Recog.})

@String(ICCV= {Int. Conf. Comput. Vis.})

@String(ECCV= {Eur. Conf. Comput. Vis.})

@String(ICLR = {Int. Conf. Learn. Represent.})

@String(CVPR  = {CVPR})

@String(ICCV  = {ICCV})

@String(ECCV  = {ECCV})

@String(ICLR  = {ICLR})

@String(CVPR  = {IEEE Conf. Comput. Vis. Pattern Recog.})

@String(ICCV  = {Int. Conf. Comput. Vis.})

@String(ECCV  = {Eur. Conf. Comput. Vis.})

@String(NeurIPS = {Adv. Neural Inform. Process. Syst.})

@String(ICLR  = {Int. Conf. Learn. Represent.})

@String(NeurIPS = {NeurIPS})

@inproceedings{park2022majority,
  title={The majority can help the minority: Context-rich minority oversampling for long-tailed classification},
  author={Park, Seulki and Hong, Youngkyu and Heo, Byeongho and Yun, Sangdoo and Choi, Jin Young},
  booktitle={Proceedings of the IEEE/CVF conference on computer vision and pattern recognition},
  pages={6887--6896},
  year={2022}
}

@article{cao2019learning,
  title={Learning imbalanced datasets with label-distribution-aware margin loss},
  author={Cao, Kaidi and Wei, Colin and Gaidon, Adrien and Arechiga, Nikos and Ma, Tengyu},
  journal={Advances in neural information processing systems},
  volume={32},
  year={2019}
}

@inproceedings{wang2024enhance,
  title={Enhance image classification via inter-class image mixup with diffusion model},
  author={Wang, Zhicai and Wei, Longhui and Wang, Tan and Chen, Heyu and Hao, Yanbin and Wang, Xiang and He, Xiangnan and Tian, Qi},
  booktitle={Proceedings of the IEEE/CVF conference on computer vision and pattern recognition},
  pages={17223--17233},
  year={2024}
}

@inproceedings{koh2025synthetic,
  title={Synthetic Data Augmentation using Pre-trained Diffusion Models for Long-tailed Food Image Classification},
  author={Koh, GaYeon and Oh, Hyun-Jic and Noh, Jeonghyun and Jeong, Won-Ki},
  booktitle={Proceedings of the Computer Vision and Pattern Recognition Conference},
  pages={391--400},
  year={2025}
}

@article{ye2023synaug,
  title={{SYNAuG}: Exploiting Synthetic Data for Data Imbalance Problems},
  author={Ye-Bin, Moon and Hyeon-Woo, Nam and Choi, Wonseok and Kim, Nayeong and Kwak, Suha and Oh, Tae-Hyun},
  journal={arXiv preprint arXiv:2308.00994},
  year={2023}
}

@article{zhang2017mixup,
  title={mixup: Beyond empirical risk minimization},
  author={Zhang, Hongyi and Cisse, Moustapha and Dauphin, Yann N and Lopez-Paz, David},
  journal={arXiv preprint arXiv:1710.09412},
  year={2017}
}

@inproceedings{rombach2022high,
  title={High-resolution image synthesis with latent diffusion models},
  author={Rombach, Robin and Blattmann, Andreas and Lorenz, Dominik and Esser, Patrick and Ommer, Bj{\"o}rn},
  booktitle=CVPR,
  pages={10684--10695},
  year={2022}
}

@inproceedings{yun2019cutmix,
  title={Cutmix: Regularization strategy to train strong classifiers with localizable features},
  author={Yun, Sangdoo and Han, Dongyoon and Oh, Seong Joon and Chun, Sanghyuk and Choe, Junsuk and Yoo, Youngjoon},
  booktitle=ICCV,
  pages={6023--6032},
  year={2019}
}

@inproceedings{wang2025inversion,
  title={Inversion Circle Interpolation: Diffusion-based Image Augmentation for Data-scarce Classification},
  author={Wang, Yanghao and Chen, Long},
  booktitle={Proceedings of the Computer Vision and Pattern Recognition Conference},
  pages={25560--25569},
  year={2025}
}

@inproceedings{
  trabucco2024effective,
  title={Effective Data Augmentation With Diffusion Models},
  author={Brandon Trabucco and Kyle Doherty and Max Gurinas and Ruslan Salakhutdinov},
  booktitle={The Twelfth International Conference on Learning Representations},
  year={2024},
  url={https://openreview.net/forum?id=ZWzUA9zeAg}
}

@article{he2022synthetic,
  title={Is synthetic data from generative models ready for image recognition?},
  author={He, Ruifei and Sun, Shuyang and Yu, Xin and Xue, Chuhui and Zhang, Wenqing and Torr, Philip and Bai, Song and Qi, Xiaojuan},
  journal={arXiv preprint arXiv:2210.07574},
  year={2022}
}

@inproceedings{hu2022lora,
  author    = {Edward J. Hu and
               Yelong Shen and
               Phillip Wallis and
               Zeyuan Allen-Zhu and
               Yuanzhi Li and
               Shean Wang and
               Lu Wang and
               Weizhu Chen},
  title     = {LoRA: Low-Rank Adaptation of Large Language Models},
  booktitle = {ICLR},
  year      = {2022},
  url       = {https://openreview.net/forum?id=nZeVKeeFYf9}
}

@inproceedings{cao2023masactrl,
  title={Masactrl: Tuning-free mutual self-attention control for consistent image synthesis and editing},
  author={Cao, Mingdeng and Wang, Xintao and Qi, Zhongang and Shan, Ying and Qie, Xiaohu and Zheng, Yinqiang},
  booktitle={Proceedings of the IEEE/CVF international conference on computer vision},
  pages={22560--22570},
  year={2023}
}

@article{jeong2024visual,
  title={Visual style prompting with swapping self-attention},
  author={Jeong, Jaeseok and Kim, Junho and Choi, Yunjey and Lee, Gayoung and Uh, Youngjung},
  journal={arXiv preprint arXiv:2402.12974},
  year={2024}
}

@inproceedings{gal2023textual,
  title={An Image is Worth One Word: Personalizing Text-to-Image Generation using Textual Inversion},
  author={Gal, Rinon and Alaluf, Yuval and Atzmon, Yuval and Patashnik, Or and Bermano, Amit H. and Chechik, Gal and Cohen-Or, Daniel},
  booktitle={The Eleventh International Conference on Learning Representations},
  year={2023},
  url={https://openreview.net/forum?id=NAQvF08TcyG}
}

@inproceedings{luddecke2022image,
  title={Image segmentation using text and image prompts},
  author={L{\"u}ddecke, Timo and Ecker, Alexander},
  booktitle={Proceedings of the IEEE/CVF conference on computer vision and pattern recognition},
  pages={7086--7096},
  year={2022}
}

@article{ho2020denoising,
  title={Denoising diffusion probabilistic models},
  author={Ho, Jonathan and Jain, Ajay and Abbeel, Pieter},
  journal={Advances in neural information processing systems},
  volume={33},
  pages={6840--6851},
  year={2020}
}

@article{song2020denoising,
  title={Denoising diffusion implicit models},
  author={Song, Jiaming and Meng, Chenlin and Ermon, Stefano},
  journal={arXiv preprint arXiv:2010.02502},
  year={2020}
}

@article{zhang2023deep,
  title={Deep long-tailed learning: A survey},
  author={Zhang, Yifan and Kang, Bingyi and Hooi, Bryan and Yan, Shuicheng and Feng, Jiashi},
  journal={IEEE transactions on pattern analysis and machine intelligence},
  volume={45},
  number={9},
  pages={10795--10816},
  year={2023},
  publisher={IEEE}
}

@article{dosovitskiy2020vit,
  title={An Image is Worth 16x16 Words: Transformers for Image Recognition at Scale},
  author={Dosovitskiy, Alexey and Beyer, Lucas and Kolesnikov, Alexander and Weissenborn, Dirk and Zhai, Xiaohua and Unterthiner, Thomas and  Dehghani, Mostafa and Minderer, Matthias and Heigold, Georg and Gelly, Sylvain and Uszkoreit, Jakob and Houlsby, Neil},
  journal={ICLR},
  year={2021}
}

@article{azizi2023synthetic,
  title={Synthetic Data from Diffusion Models Improves ImageNet Classification},
  author={Azizi, Shekoofeh and Kornblith, Simon and Saharia, Chitwan and Norouzi, Mohammad and Fleet, David J.},
  journal={Transactions on Machine Learning Research},
  issn={2835-8856},
  year={2023},
  url={https://openreview.net/forum?id=DlRsoxjyPm}
}

@article{xu2025deep,
  title={Deep learning for cross-domain few-shot visual recognition: A survey},
  author={Xu, Huali and Zhi, Shuaifeng and Sun, Shuzhou and Patel, Vishal and Liu, Li},
  journal={ACM Computing Surveys},
  volume={57},
  number={8},
  pages={1--37},
  year={2025},
  publisher={ACM New York, NY}
}

@article{miyai2024generalized,
  title={Generalized out-of-distribution detection and beyond in vision language model era: A survey},
  author={Miyai, Atsuyuki and Yang, Jingkang and Zhang, Jingyang and Ming, Yifei and Lin, Yueqian and Yu, Qing and Irie, Go and Joty, Shafiq and Li, Yixuan and Li, Hai and others},
  journal={arXiv preprint arXiv:2407.21794},
  year={2024}
}

@inproceedings{samuel2021generalized,
  title={From generalized zero-shot learning to long-tail with class descriptors},
  author={Samuel, Dvir and Atzmon, Yuval and Chechik, Gal},
  booktitle={Proceedings of the IEEE/CVF winter conference on applications of computer vision},
  pages={286--295},
  year={2021}
}

@inproceedings{he2016deep,
  title={Deep residual learning for image recognition},
  author={He, Kaiming and Zhang, Xiangyu and Ren, Shaoqing and Sun, Jian},
  booktitle={Proceedings of the IEEE conference on computer vision and pattern recognition},
  pages={770--778},
  year={2016}
}

@article{wah2011caltech,
  title={The caltech-ucsd birds-200-2011 dataset},
  author={Wah, Catherine and Branson, Steve and Welinder, Peter and Perona, Pietro and Belongie, Serge},
  year={2011},
  publisher={california institute of technology}
}

@inproceedings{parkhi2012cats,
  title={Cats and dogs},
  author={Parkhi, Omkar M and Vedaldi, Andrea and Zisserman, Andrew and Jawahar, CV},
  booktitle={2012 IEEE conference on computer vision and pattern recognition},
  pages={3498--3505},
  year={2012},
  organization={IEEE}
}

@article{maji2013fine,
  title={Fine-grained visual classification of aircraft},
  author={Maji, Subhransu and Rahtu, Esa and Kannala, Juho and Blaschko, Matthew and Vedaldi, Andrea},
  journal={arXiv preprint arXiv:1306.5151},
  year={2013}
}

@inproceedings{krause20133d,
  title={3d object representations for fine-grained categorization},
  author={Krause, Jonathan and Stark, Michael and Deng, Jia and Fei-Fei, Li},
  booktitle={Proceedings of the IEEE international conference on computer vision workshops},
  pages={554--561},
  year={2013}
}

@article{sagawa2019distributionally,
  title={Distributionally robust neural networks for group shifts: On the importance of regularization for worst-case generalization},
  author={Sagawa, Shiori and Koh, Pang Wei and Hashimoto, Tatsunori B and Liang, Percy},
  journal={arXiv preprint arXiv:1911.08731},
  year={2019}
}

@inproceedings{hertz2023prompt,
  title     = {Prompt-to-Prompt Image Editing with Cross Attention Control},
  author    = {Amir Hertz and Ron Mokady and Jay Tenenbaum and Kfir Aberman and Yael Pritch and Daniel Cohen-Or},
  booktitle = {The 11th International Conference on Learning Representations (ICLR)},
  year      = {2023},
  url       = {https://openreview.net/forum?id=_CDixzkzeyb}
}

@inproceedings{kirillov2023segment,
  title={Segment anything},
  author={Kirillov, Alexander and others},
  booktitle={Proc. ICCV},
  year={2023}
}

@inproceedings{gu2023photoswap,
  title={Photoswap: Personalized subject swapping in images},
  author={Gu, Jing and others},
  booktitle={Proc. NeurIPS},
  year={2023}
}

@inproceedings{chen2024anydoor,
  title={Anydoor: Zero-shot object-level image customization},
  author={Chen, Xi and others},
  booktitle={Proc. CVPR},
  year={2024}
}

@article{li2025domain,
  title={Domain-RAG: Retrieval-Guided Compositional Image Generation for Cross-Domain Few-Shot Object Detection},
  author={Li, Yu and others},
  journal={arXiv:2506.05872},
  year={2025}
}

@inproceedings{xie2023boxdiff,
  title={Boxdiff: Text-to-image synthesis with training-free box-constrained diffusion},
  author={Xie, Jinheng and others},
  booktitle={Proc. ICCV},
  year={2023}
}

@inproceedings{li2024simple,
  title={A simple background augmentation method for object detection with diffusion model},
  author={Li, Yuhang and others},
  booktitle={Proc. ECCV},
  year={2024}
}

@article{lu2025higfa,
  title={HiGFA: Hierarchical Guidance for Fine-grained Data Augmentation with Diffusion Models},
  author={Lu, Zhiguang and Xu, Qianqian and Wen, Peisong and Dai, Siran and Huang, Qingming},
  journal={arXiv preprint arXiv:2511.12547},
  year={2025}
}

@article{zhao2025salient,
  title={Salient Concept-Aware Generative Data Augmentation},
  author={Zhao, Tianchen and Chen, Xuanbai and Li, Zhihua and Fang, Jun and An, Dongsheng and Xu, Xiang and Tu, Zhuowen and Xing, Yifan},
  journal={arXiv preprint arXiv:2510.15194},
  year={2025}
}

@article{zhu2025recon,
  title={ReCon: Region-Controllable Data Augmentation with Rectification and Alignment for Object Detection},
  author={Zhu, Haowei and Pan, Tianxiang and Qin, Rui and Yong, Jun-Hai and Wang, Bin},
  journal={arXiv preprint arXiv:2510.15783},
  year={2025}
}

@inproceedings{yang2026beyond,
  title={Beyond Objects: Contextual Synthetic Data Generation for Fine-Grained Classification},
  author={Yang, William and Wu, Xindi and Deng, Zhiwei and Tureci, Esin and Russakovsky, Olga},
  booktitle={Proceedings of the IEEE/CVF Conference on Computer Vision and Pattern Recognition},
  pages={22498--22508},
  year={2026}
}

@article{nichol2021glide,
  title={Glide: Towards photorealistic image generation and editing with text-guided diffusion models},
  author={Nichol, Alex and Dhariwal, Prafulla and Ramesh, Aditya and Shyam, Pranav and Mishkin, Pamela and McGrew, Bob and Sutskever, Ilya and Chen, Mark},
  journal={arXiv preprint arXiv:2112.10741},
  year={2021}
}

@article{kim2025diverse,
  title={Diverse Text-to-Image Generation via Contrastive Noise Optimization},
  author={Kim, Byungjun and Um, Soobin and Ye, Jong Chul},
  journal={arXiv preprint arXiv:2510.03813},
  year={2025}
}

@inproceedings{sadat2024cads,
  title={CADS: Unleashing the diversity of diffusion models through condition-annealed sampling},
  author={Sadat, Seyedmorteza and Buhmann, Jakob and Bradley, Derek and Hilliges, Otmar and Weber, Romann},
  booktitle={International Conference on Learning Representations},
  volume={2024},
  pages={23723--23755},
  year={2024}
}

@inproceedings{guo2024initno,
  title={Initno: Boosting text-to-image diffusion models via initial noise optimization},
  author={Guo, Xiefan and Liu, Jinlin and Cui, Miaomiao and Li, Jiankai and Yang, Hongyu and Huang, Di},
  booktitle={Proceedings of the IEEE/CVF Conference on Computer Vision and Pattern Recognition},
  pages={9380--9389},
  year={2024}
}

@article{carion2025sam,
  title={Sam 3: Segment anything with concepts},
  author={Carion, Nicolas and Gustafson, Laura and Hu, Yuan-Ting and Debnath, Shoubhik and Hu, Ronghang and Suris, Didac and Ryali, Chaitanya and Alwala, Kalyan Vasudev and Khedr, Haitham and Huang, Andrew and others},
  journal={arXiv preprint arXiv:2511.16719},
  year={2025}
}

@inproceedings{harrington2026s,
  title={It's Never Too Late: Noise Optimization for Collapse Recovery in Trained Diffusion Models},
  author={Harrington, Anne and Koepke, A and Karthik, Shyamgopal and Darrell, Trevor and Efros, Alexei A},
  booktitle={Proceedings of the IEEE/CVF Conference on Computer Vision and Pattern Recognition},
  pages={43124--43134},
  year={2026}
}

@article{oquab2023dinov2,
  title={Dinov2: Learning robust visual features without supervision},
  author={Oquab, Maxime and Darcet, Timoth{\'e}e and Moutakanni, Th{\'e}o and Vo, Huy and Szafraniec, Marc and Khalidov, Vasil and Fernandez, Pierre and Haziza, Daniel and Massa, Francisco and El-Nouby, Alaaeldin and others},
  journal={arXiv preprint arXiv:2304.07193},
  year={2023}
}

@article{friedman2023vendi,
  title={The Vendi Score: A Diversity Evaluation Metric for Machine Learning},
  author={Friedman, Dan and Dieng, Adji Bousso},
  journal={Transactions on Machine Learning Research},
  issn={2835-8856},
  year={2023}
}

@inproceedings{oh2024controllable,
  title={Controllable and efficient multi-class pathology nuclei data augmentation using text-conditioned diffusion models},
  author={Oh, Hyun-Jic and Jeong, Won-Ki},
  booktitle={International Conference on Medical Image Computing and Computer-Assisted Intervention},
  pages={36--46},
  year={2024},
  organization={Springer}
}
}






\clearpage
\twocolumn[{%
\begin{center}
    {\Large \bfseries
    Improving Diffusion-Based Augmentation for Data-Scarce Image Classification via Mask-Guided Reference Attention and Repulsive Noise Optimization
    \par}
    {\large Supplementary Materials\par}
\end{center}
}]

\setcounter{section}{0}
\renewcommand{\thesection}{S\arabic{section}}
\renewcommand{\thesubsection}{\thesection.\arabic{subsection}}

In this supplementary material, we provide additional implementation details and analyses complementing the main paper, organized as follows:
\begin{itemize}
    \item Implementation details (\cref{sec:impl})
    \item Sampling algorithm (\cref{sec:alg})
    \item MGR Attention analysis (\cref{sec:mgr_analysis})
    \item Segmentation backbone sensitivity (\cref{sec:seg})
    \item RNO hyperparameter sensitivity (\cref{sec:rno_sens})
    \item Comparison to Reference-Guided Methods (\cref{sec:ref_guided})
    \item Additional qualitative results (\cref{sec:qual})
\end{itemize}









\section{Implementation Details}
\label{sec:impl}
\paragraph{Reference mask extraction.}
Reference masks are pre-computed offline using SAM3.
To ensure mask validity, each image--mask overlay is automatically checked using Qwen2.5-VL-3B-Instruct\footnote{
\url{https://huggingface.co/Qwen/Qwen2.5-VL-3B-Instruct}}. 
Among 16,890 reference images across all datasets, 36 (0.21\%) produced invalid masks and were replaced with revised masks.

\paragraph{RNO hyperparameters.}
RNO uses 3 optimization steps with learning rate (LR) 0.01 for CUB-LT, 10 steps with LR 0.04 for Aircraft, Cars, and Pet, and 30 steps with LR 0.08 for Flower-LT.
We use $B{=}5$ initial noises per optimization batch; each noise is optimized as a target, with the remaining $B{-}1{=}4$ noises conditioned on $\mathbf{c}_a$ serving as its frozen (stop-gradient) adaptation-prompt anchors (Eq.~\ref{eq:rno_loss}).


\section{Sampling Algorithm}
\label{sec:alg}
Algorithm~\ref{alg:method} summarizes the proposed sampling process.
RNO optimizes target initial noises against frozen adaptation-prompt anchors before denoising begins (lines 4--11).
The denoising then proceeds from $t{=}T$ to $t{=}t_m$ (lines 19--20) to establish spatial structure for mask extraction (lines 14--16), after which MGR Attention is activated for the remaining steps ($t_m{\to}0$, lines 17--18).

\begin{algorithm}[t]
\small
\caption{Sampling with MGR Attention and RNO}
\label{alg:method}
\begin{algorithmic}[1]
\State \textbf{Input:} adapted denoiser $\boldsymbol{\epsilon}_{\theta}$, adaptation prompt $\mathbf{c}_a$, target prompts $\{\mathbf{c}^i_t\}_{i=1}^{B}$, reference image $\mathbf{I}_r$, mask timestep $t_m$, $T{=}50$, RNO steps $K$, LR $\eta$
\State \textit{// $\mathbf{M}_r$: pre-computed from $\mathbf{I}_r$ via SAM3 (offline)}
\State Initialize $\{\mathbf{z}_T^i\}_{i=1}^B \sim \mathcal{N}(\mathbf{0}, \mathbf{I})$
\State \textit{// RNO: optimize target noises}
\State Compute $\hat{\mathbf{z}}^j_a {=} \mathrm{sg}[\hat{\mathbf{z}}_{0|T} (\mathbf{z}_T^j, \mathbf{c}_a)]$ for all $j$
\For{$k = 1$ \textbf{to} $K$}
    \State
    Compute $\hat{\mathbf{z}}^i {=} \hat{\mathbf{z}}_{0|T}(\mathbf{z}_T^i, \mathbf{c}_t^i)$ for all $i$
    \State
    $\mathcal{L}_{RNO} {=} \frac{1}{B(B{-}1)}
    \sum_i\sum_{j\neq i} 
    \mathrm{sim}(\hat{\mathbf{z}}^i, 
    \hat{\mathbf{z}}^j_a)$ (\Cref{eq:rno_loss})
    \State
    $\mathbf{z}_T^i \leftarrow 
    \mathbf{z}_T^i - \eta 
    \nabla_{\mathbf{z}_T^i} \mathcal{L}_{RNO}$
    for all $i$
\EndFor
\State $\{\mathbf{z}_T^{i*}\}_{i=1}^B \leftarrow \{\mathbf{z}_T^i\}_{i=1}^B$ \textit{// optimized noises}
\State \textit{// Denoising: $T \to 0$}
\For{$t = T$ \textbf{downto} $1$}
    \If{$t = t_m$}
        \State
        $\{\mathbf{M}_t^i\} \leftarrow 
        \mathrm{SAM3}(\mathrm{decode}
        (\hat{\mathbf{z}}_{0|t_m}^{i*}))$
        for all $i$
    \EndIf
    \If{$t \leq t_m$}
        \State
        $\mathbf{z}_{t-1}^{i*} \leftarrow 
        \mathrm{DDIM\_with\_MGR}(\mathbf{z}_t^{i*}, 
        \mathbf{c}_t^i, \mathbf{M}_t^i, 
        \mathbf{M}_r, \mathbf{I}_r)$
    \Else
        \State
        $\mathbf{z}_{t-1}^{i*} \leftarrow 
        \mathrm{DDIM}(\mathbf{z}_t^{i*}, 
        \mathbf{c}_t^i)$
    \EndIf
\EndFor
\State \Return $\{\mathrm{decode} (\hat{\mathbf{z}}_0^{i*})\}_{i=1}^B$
\end{algorithmic}
\end{algorithm}

\begin{figure}[t!]
    \centering
    \includegraphics[width=1.02\linewidth]{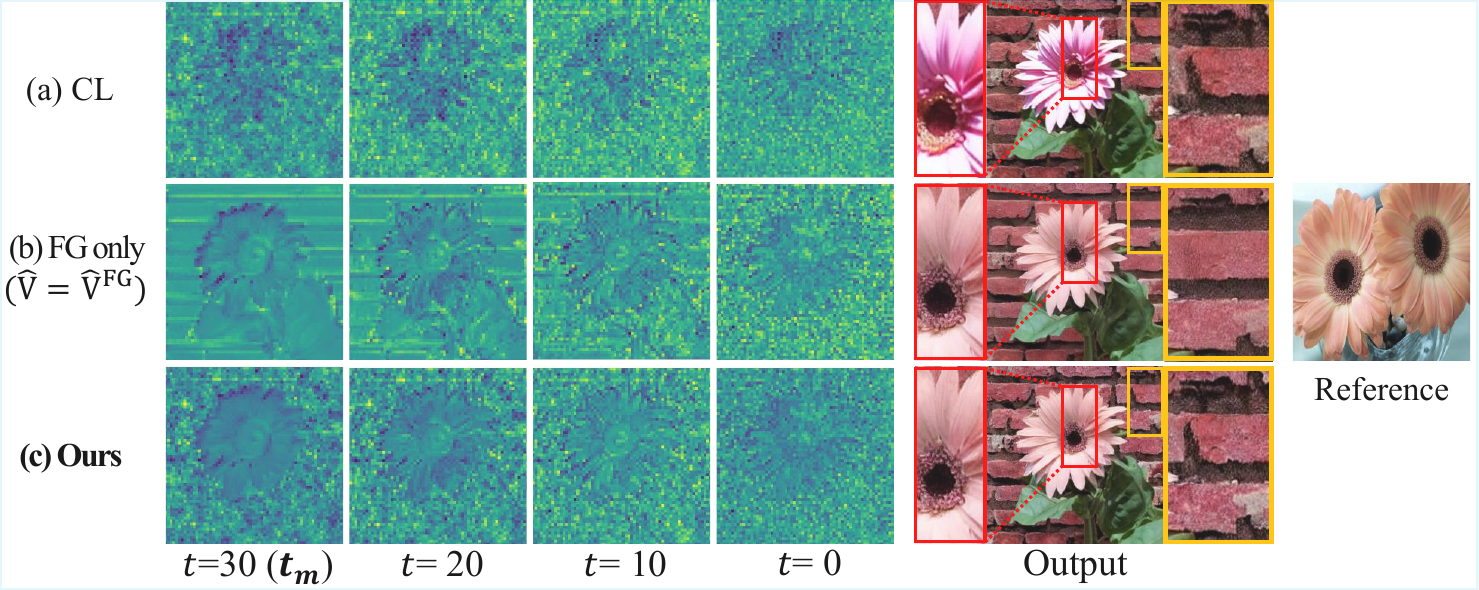}
    \caption{
    Target value features ($\hat{\mathbf{V}}$) over denoising timesteps and final outputs:
    (a) CL, (b) foreground-only update ($\hat{\mathbf{V}}{=}\hat{\mathbf{V}}^{\mathrm{FG}}$), and (c) MGR Attention.
    MGR aligns foreground activations with the reference while keeping target responses outside the foreground.}
    \label{fig:attn_vis}
\end{figure}

\begin{figure}[t!]
    \centering
    \includegraphics[width=1.0\linewidth]{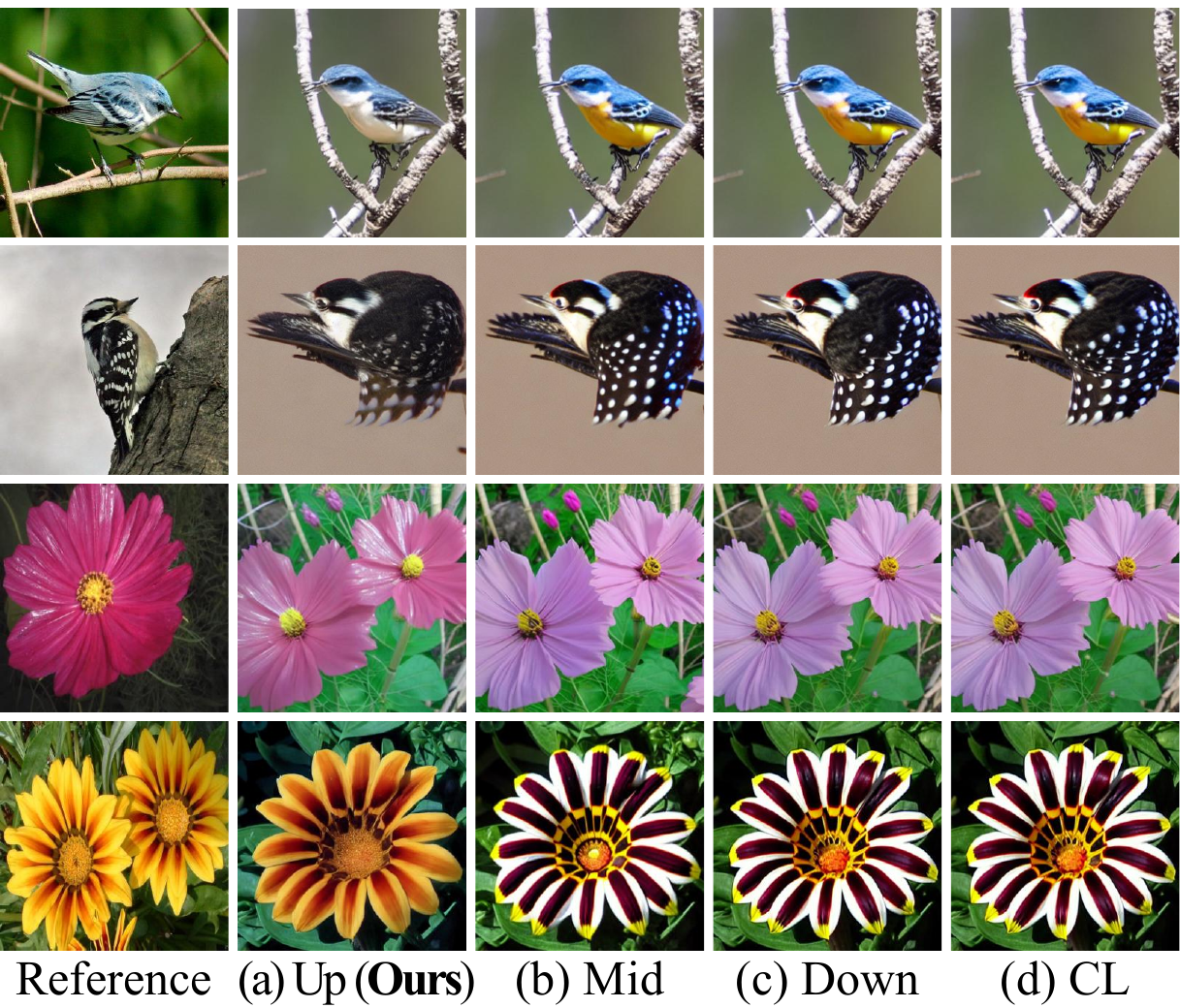}
    \caption{
    Layer selection for MGR Attention on CUB-LT IF100.
    (a) Up-blocks (Ours), (b) Mid-blocks, (c) Down-blocks, (d) CL baseline.
    Up-blocks best preserve foreground details. 
    }
    \label{fig:layer_mgr}
\end{figure}

\section{Analysis on MGR Attention}
\label{sec:mgr_analysis}

\paragraph{Value feature visualization.}
\Cref{fig:attn_vis} visualizes target value features $\hat{\mathbf{V}}$ during denoising across timesteps, depicting the effect of the value update in~\cref{eq:mgr_merge}.
CL activations (a) remain dominated by the adapted class prior throughout denoising.
Foreground-only update (b) produces more structured foreground responses, but background regions appear smoothed in the output image.
MGR Attention (c, Ours) strengthens foreground responses aligned with the reference while retaining target background activations, producing more structured non-foreground responses than (b).
These results support retaining target background values in the value update, as formulated in~\cref{eq:mgr_merge}.


\paragraph{Layer selection.}
\Cref{fig:layer_mgr,tab:layer_selection} ablate the self-attention layer selection for MGR Attention on CUB-LT IF100.
In~\cref{tab:layer_selection}, decoder up-blocks consistently yield the best accuracy; we apply MGR Attention to up-block layers from layer 2.1, selected empirically.
Mid-blocks show slightly higher Medium accuracy but underperform on tail classes, while down-blocks perform worst overall, producing outputs closer to the CL baseline as shown in~\cref{fig:layer_mgr}.

\begin{figure}
    \centering
    \includegraphics[width=1.0\linewidth]{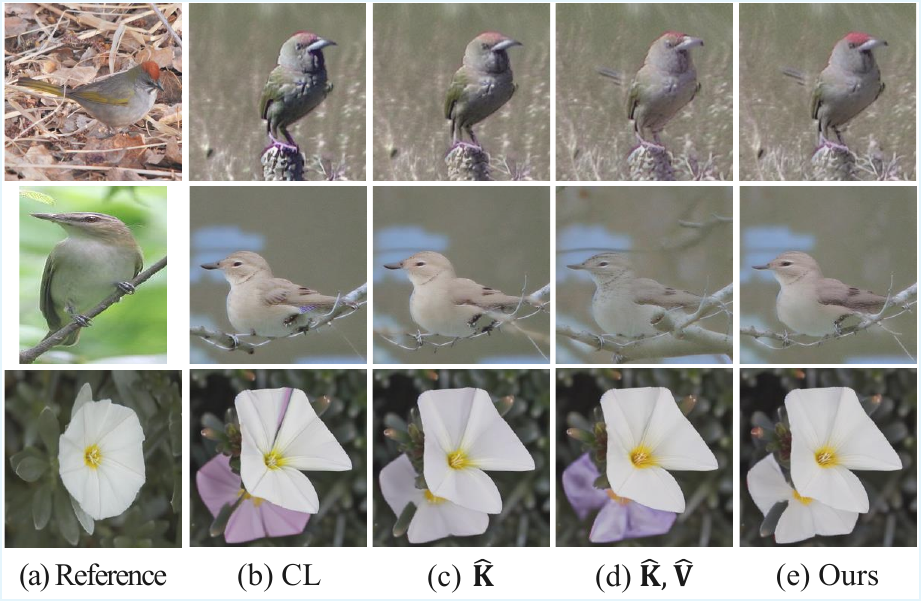}
    \caption{
    Feature modification strategy on CUB-LT IF100 and Flower-LT IF100. 
    (a) Reference, (b) CL baseline, (c) key-only ($\hat{\mathbf{K}}$), (d) key-value ($\hat{\mathbf{K}}, \hat{\mathbf{V}}$), (e) Ours: foreground value update ($\hat{\mathbf{V}}$).
    }
    \label{fig:selective_value_mod}
\end{figure}

\paragraph{Feature modification strategy.}
\Cref{fig:selective_value_mod} provides visual evidence for the foreground value update design in MGR Attention. 
Our foreground-selective value update (e, $\hat{\mathbf{V}}$) empirically yields a better balance between reference guidance and layout preservation across both datasets.


\begin{table}[t]
\centering
\small
\renewcommand{\arraystretch}{1.0}
\setlength{\tabcolsep}{8pt}
\caption{
Layer selection analysis for MGR Attention on CUB-LT IF100 (ResNet-50).
Decoder up-blocks yield the best performance. 
}
\resizebox{\linewidth}{!}{
\begin{tabular}{lccccc}
\toprule
Method & Many & Medium & Few & Single & Avg \\
\midrule
Down-blocks 
        & 82.89 & 70.48 & 42.01 & 36.28 & 53.62 \\
Mid-blocks  
        & 84.82 & \textbf{71.10} & 42.95 
        & 37.55 & 54.54 \\
Up-blocks (Ours) 
        & \textbf{85.54} & 70.87 
        & \textbf{44.70} & \textbf{39.46} 
        & \textbf{55.61} \\
\bottomrule
\end{tabular}
}
\label{tab:layer_selection}
\end{table}

\begin{figure}
    \centering
    \includegraphics[width=1.0\linewidth]{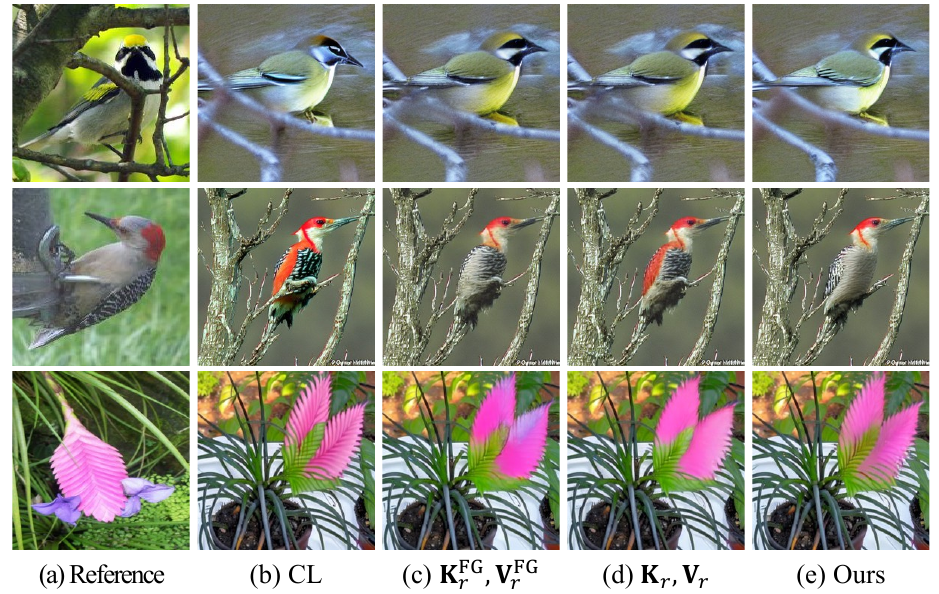}
    \caption{
    KV bank composition on CUB-LT and Flower-LT IF100. 
    (a) Reference, (b) CL baseline, (c) foreground-only ($\mathbf{K}_r^{\mathrm{FG}}, \mathbf{V}_r^{\mathrm{FG}}$), (d) reference-full ($\mathbf{K}_r, \mathbf{V}_r$), (e) Ours ($\mathbf{K}_{\mathrm{bank}} {=} [\mathbf{K}^{\mathrm{BG}}; \mathbf{K}_r^{\mathrm{FG}}]$).
    }
    \label{fig:supple_attn_bank}
\end{figure}

\paragraph{Attention bank composition.}
\Cref{fig:supple_attn_bank} provides visual evidence for the KV bank design in MGR Attention.
Using foreground-only reference tokens (c, $\mathbf{K}_r^{\mathrm{FG}}, \mathbf{V}_r^{\mathrm{FG}}$) or using full reference tokens (d, $\mathbf{K}_r, \mathbf{V}_r$) can produce incomplete structures (row 2).
Our bank design (e, $\mathbf{K}_{\mathrm{bank}} {=} [\mathbf{K}^{\mathrm{BG}}; \mathbf{K}_r^{\mathrm{FG}}]$) empirically yields a better balance between foreground guidance and spatial grounding.


\begin{table} 
\centering
\caption{
Segmentation backbone analysis on CUB-LT IF100 
(MGR+RNO, ResNet-50).
}
\renewcommand{\arraystretch}{1.0}
\resizebox{\linewidth}{!}{
\begin{tabular}{lccccc}
\toprule
Segmenter & Many & Medium & Few & Single & Avg \\
\midrule
CLIPSeg~\cite{luddecke2022image}
        & 84.34 & 70.19 & 45.80 & 40.44 & 56.01 \\
SAM3~\cite{carion2025sam}\checkmark
        & 85.78 & 69.57 & 46.08 & 41.05 & 56.09 \\
\bottomrule
\end{tabular}
}
\label{tab:seg_backbone}
\end{table}


\begin{figure*}[t!]
\centering
\includegraphics[width=.9\linewidth]{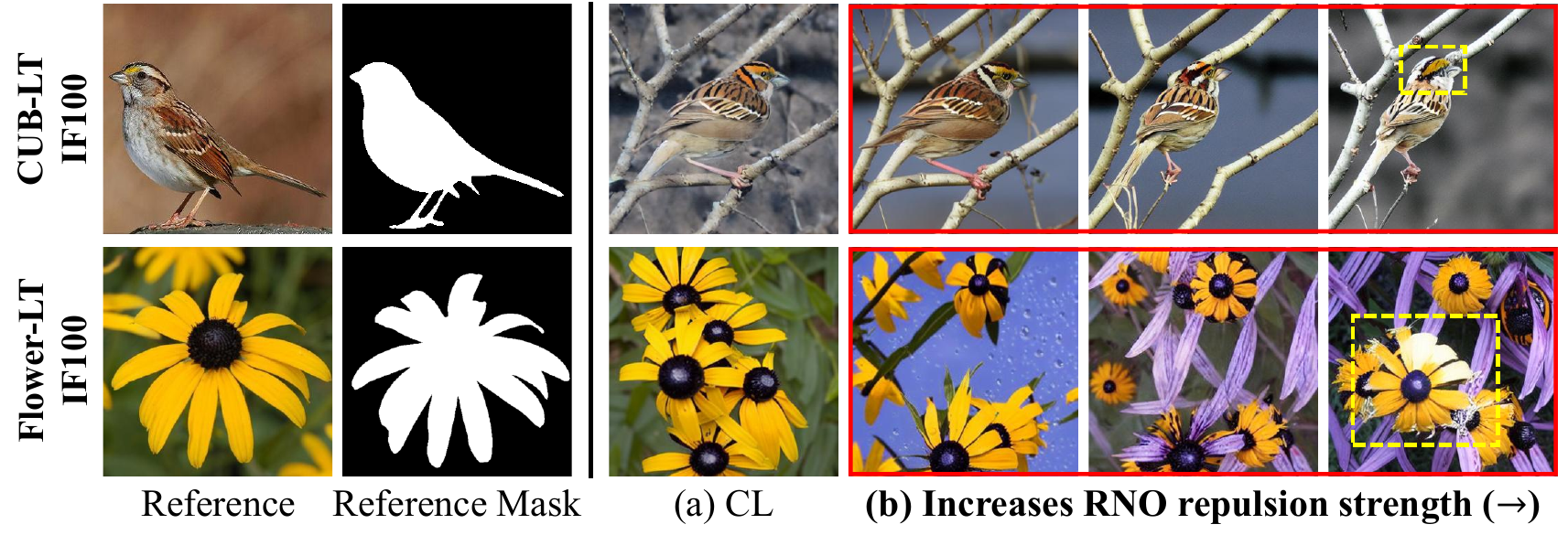}
\caption{
Effect of increasing RNO repulsion strength on CUB-LT and Flower-LT IF100.
(a) CL.
(b) Ours with increasing RNO repulsion strength from left to right, starting from the default setting (CUB: 3 steps, LR 0.01; Flower: 30 steps, LR 0.08). Prompts: ``sitting on a branch of a tree'' (CUB) and ``with water droplets on it'' (Flower).
Greater repulsion introduces layout diversity but risks drifting from class-consistent foreground appearance (yellow boxes).
}
\label{fig:rno_sensitivity}
\end{figure*}

\section{Segmentation Backbone}
\label{sec:seg}
\Cref{tab:seg_backbone} compares CLIPSeg~\cite{luddecke2022image} and SAM3~\cite{carion2025sam} as zero-shot segmenters on CUB-LT IF100.
Both backbones show comparable overall performance (Avg within 0.08 points), indicating that MGR Attention is robust to the choice of segmentation backbone.
SAM3 achieves slightly better Few and overall accuracy, and is used as the default segmenter in all main paper experiments.


\section{RNO Sensitivity}
\label{sec:rno_sens}
\Cref{fig:rno_sensitivity} illustrates the effect of increasing repulsion strength on generated samples.
With the default setting (leftmost column of (b)), MGR+RNO preserves foreground details while generating diverse layouts.
As repulsion strength increases (remaining columns of (b)), generated samples show greater layout variation but can drift from class-consistent foreground appearance, suggesting that applying too strong a repulsion objective may push latent trajectories away from class-consistent regions.

\begin{table}[t]
\centering
\caption{
RNO hyperparameter sensitivity on CUB-LT IF100 (MGR+RNO, ResNet-50).
}
\renewcommand{\arraystretch}{.9}
\resizebox{\linewidth}{!}{
\begin{tabular}{llccccc}
\toprule
Steps & LR & Many & Medium & Few & Single & Avg \\
\midrule
30 & 0.08 
        & 84.34 & 70.48 & 44.86 & 39.22 & 55.51 \\
10 & 0.04 
        & 85.39 & 70.66 & 44.36 & 39.31 & 55.67 \\
5  & 0.04 
        & 84.82 & \textbf{71.61} & 45.14 & 39.87 & 56.06 \\
3  & 0.01\checkmark 
        & \textbf{85.78} & 69.57 
        & \textbf{46.08} & \textbf{41.05} 
        & \textbf{56.09} \\
\bottomrule
\end{tabular}
}
\label{tab:rno_sensitivity}
\end{table}


\Cref{tab:rno_sensitivity} ablates RNO hyperparameter sensitivity on CUB-LT IF100.
The empirically selected setting (3 steps, LR 0.01) achieves the best Few, Single, and overall accuracy, while stronger repulsion settings do not consistently improve performance across all groups.


\section{Comparison to Reference-Guided Methods}
\label{sec:ref_guided}
\Cref{tab:attn_comparison} compares MGR Attention against attention-based editing and detection-oriented augmentation methods on CUB-LT IF100. 
Attention editing methods (P2P~\cite{hertz2023prompt}, PHOTOSWAP~\cite{gu2023photoswap}, MasaCtrl~\cite{cao2023masactrl}, VSP~\cite{jeong2024visual}) show limited gains on Few and Single groups as global feature manipulation degrades foreground details, while detection-oriented methods (Li~\etal~\cite{li2024simple}, Domain-RAG~\cite{li2025domain}) require bounding box annotations or object-level objectives and underperform on tail classes in our classification setting.
MGR consistently improves Few and Single accuracy, and MGR+RNO achieves the best overall performance.

\begin{table} 
\caption{
Comparison with attention-based editing and detection-oriented augmentation on CUB-LT IF100 (ResNet-50).
All rows are our runs under the same protocol (main-paper Tab.~\ref{tab:lt_results} reports CE from Diff-II~\cite{wang2025inversion});
attention-editing baselines use their original settings, unlike the in-pipeline K/V variants in main-paper Tab.~\ref{tab:design_choices}.
Groups (classes): Many $x{>}20$ (16), Medium $5{\leq}x{\leq}20$ (61), Few $x{<}5$ incl.\ Single (123), Single $x{=}1$ (83).
}
\centering
\renewcommand{\arraystretch}{0.95}
\resizebox{\linewidth}{!}{
\begin{tabular}{lcccccc}
\toprule
Method & Many & Medium & Few & Single & Avg \\
\midrule
CE only & 85.54 & 67.14 & 18.33 & 7.38 & 38.04 \\
Diff-II$\dagger$~\cite{wang2025inversion} 
        & 83.86 & 70.14 & 42.01 & 36.69 & 53.59 \\
\midrule
P2P~\cite{hertz2023prompt} 
        & 84.82 & 69.91 & 41.40 & 35.55 & 53.21 \\
PHOTOSWAP~\cite{gu2023photoswap} 
        & 83.37 & 63.57 & 19.83 & 11.62 & 37.73 \\
MasaCtrl~\cite{cao2023masactrl} 
        & 84.58 & 66.74 & 18.50 & 7.75  & 37.95 \\
VSP~\cite{jeong2024visual} 
        & 85.30 & 70.81 & 42.62 & 36.77 & 54.28 \\
\midrule
Li~\etal~\cite{li2024simple} 
        & 82.17 & 64.03 & 18.64 & 10.40 & 37.04 \\
Domain-RAG~\cite{li2025domain} 
        & 84.10 & 62.67 & 19.75 & 11.29 & 37.45 \\
\midrule
\textbf{MGR (Ours)}
        & 85.54 & \textbf{70.87} 
        & 44.70 & 39.46 & 55.61 \\
\textbf{MGR+RNO (Ours)}
        & \textbf{85.78} & 69.57 
        & \textbf{46.08} & \textbf{41.05} 
        & \textbf{56.09} \\
\bottomrule
\end{tabular}
}
\label{tab:attn_comparison}
\end{table}

\section{Additional Qualitative Results}
\label{sec:qual}
\begin{figure*}[t]
    \centering
    \makebox[\textwidth][c]{%
        \includegraphics[width=0.85\textwidth]{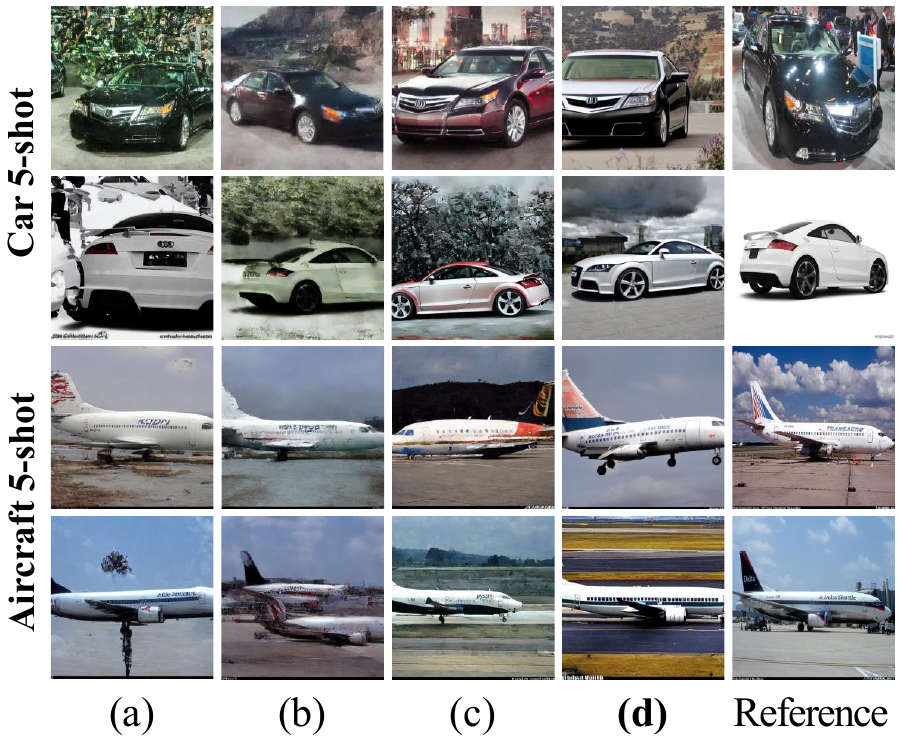}
    }
    \caption{
    Qualitative comparison on Car 5-shot and Aircraft 5-shot.
    (a) Global K/V replacement (VSP-style~\cite{jeong2024visual}),
    (b) attention-masked K/V replacement (MasaCtrl~\cite{cao2023masactrl}-style),
    (c) SAM3-masked K/V replacement, and (d) Ours (MGR Attention).
    (a) and (b) introduce global style propagation and blurry object regions;
    (c) reduces noise but can still produce blurry foreground.
    (d) better preserves target layout while enhancing localized foreground details.
    }
    \label{fig:supple_qual} 
\end{figure*}

\Cref{fig:supple_qual} visualizes the qualitative comparison to Car 5-shot and Aircraft 5-shot settings.
As in~\cref{fig:qual_module} (left), global/attention-masked K/V replacement (a, b) introduce style leakage and blurry object regions on fine-grained vehicle and aircraft textures.
SAM3-masked K/V replacement (c) reduces global noise but can still produce blurry foreground regions.
Ours (d) consistently better preserves target layout and foreground details. 


\end{document}